\documentclass[twoside,11pt]{article}

\usepackage{automl2020}
\usepackage[ruled,linesnumbered,vlined]{algorithm2e}
\usepackage{amsmath}
\usepackage{soul}
\usepackage{booktabs}

\usepackage{graphicx,subcaption}

\usepackage[svgnames]{xcolor}

\newcommand{\note}[1]{
	\noindent~\\
	\vspace{0.25cm}
	\fcolorbox{red}{Orange}{\parbox{1.0\textwidth}{#1}}
	\vspace{0.25cm}
}
\renewcommand{\note}[1]{}

\newcommand{\question}[1]{
	\noindent~\\
	\vspace{0.25cm}
	\fcolorbox{red}{green}{\parbox{1.0\textwidth}{#1}}
	\vspace{0.25cm}
}
\renewcommand{\question}[1]{}

\newcommand{\answer}[1]{
	\noindent~\\
	\vspace{0.25cm}
	\fcolorbox{red}{yellow}{\parbox{1.0\textwidth}{#1}}
	\vspace{0.25cm}
}
\renewcommand{\answer}[1]{}

\usepackage{amsmath}
\usepackage{mathtools}  % for :=
\usepackage{amsfonts}

\DeclareMathOperator*{\argmax}{arg\,max}        % Argmax
\newcommand{\emoash}{\texttt{\uppercase{sh-emoa}}{}}           
\newcommand{\mobohb}{\texttt{\uppercase{mo-bohb}}{}}
\newcommand{\mdehvi}{\texttt{\uppercase{ms-ehvi}}{}}     
\newcommand{\mobananas}{\texttt{\uppercase{mo-bananas}}{}}     
\newcommand{\bulkcut}{\texttt{\uppercase{bulk \& cut}}{}}

\newcommand{\fe}{n_{\text{fe}}}   %number of function evaluations
\newcommand{\pop}{s_{\text{pop}}}    %pupulation size
\newcommand{\cand}{\lambda_{\text{new}}}    %new candidate/configuration
\newcommand{\popul}{\mathcal{P}}
\newcommand{\pf}{pareto\_front}

\newcommand{\objectives}{f}  
\newcommand{\fcheap}{\objectives_{\text{cheap}}}    %cheap objectives
\newcommand{\fexp}{\objectives_{\text{exp}}}    %expensive
\newcommand{\surr}{\hat{\objectives}_{\text{exp}}}    %expensive
\newcommand{\neuralpred}{\hat{\objectives}}    %expensive

\jmlrheading{Julia Guerrero-Viu, Sven Hauns, Sergio Izquierdo, Guilherme Miotto, Simon Schrodi, Andr\'e Biedenkapp, Thomas Elsken, Difan Deng, Marius Lindauer and Frank Hutter}

\ShortHeadings{Multi-Objective Joint NAS and HPO}{\footnotesize{Guerrero-Viu, Hauns, Izquierdo, Miotto, Schrodi, Biedenkapp, Elsken, Deng, Lindauer, Hutter}}

\firstpageno{1}

\begin{document}

\title{Bag of Baselines for Multi-objective Joint Neural Architecture Search and Hyperparameter Optimization}

\author{\name Julia Guerrero-Viu \email guerrero@cs.uni-freiburg.de \\
      \name Sven Hauns \email haunss@tf.uni-freiburg.de \\
      \name Sergio Izquierdo \email izquierd@cs.uni-freiburg.de \\
      \name Guilherme Miotto \email alessang@cs.uni-freiburg.de \\
      \name Simon Schrodi  \email schrodi@cs.uni-freiburg.de \\
      \name Andr\'e Biedenkapp  \email biedenka@cs.uni-freiburg.de \\
      \addr University of Freiburg\\
      \name Thomas Elsken \email thomas.elsken@de.bosch.com\\
      \addr Bosch Center for Artificial Intelligence\\
      \name Difan Deng \email deng@tnt.uni-hannover.de\\
      \name Marius Lindauer \email lindauer@tnt.uni-hannover.de\\
      \addr Leibniz University Hannover\\
      \name Frank Hutter \email fh@cs.uni-freiburg.de\\
      \addr University of Freiburg and Bosch Center for Artificial Intelligence}

\maketitle

\begin{abstract}%   <- trailing '%' for backward compatibility of .sty file
Neural architecture search (NAS) and hyperparameter optimization (HPO) make deep learning accessible to non-experts by automatically finding the architecture of the deep neural network to use and tuning the hyperparameters of the used training pipeline.
While both NAS and HPO have been studied extensively in recent years, NAS methods typically assume fixed hyperparameters and vice versa --- there exists little work on joint NAS + HPO. Furthermore, NAS has recently often been framed as a multi-objective optimization problem, in order to take, e.g., resource requirements into account.
In this paper, we propose a set of methods that extend current approaches to jointly optimize neural architectures and hyperparameters with respect to multiple objectives. We hope that these methods will serve as simple baselines for future research on multi-objective joint NAS + HPO. To facilitate this, all our code is available at \url{https://github.com/automl/multi-obj-baselines}.

\end{abstract}

\section{Introduction}

Neural architecture search (NAS)~\citep{zoph-iclr17a,elsken_survey,wistuba2019survey} and hyperparameter optimization (HPO)~\citep{feurer-automlbook19a} make deep learning accessible to non-experts by automatically tuning the employed neural network architecture and the hyperparameters of a deep learning algorithm. 

While both fields have been studied extensively in recent years, NAS methods typically assume fixed hyperparameter configurations and vice versa.\footnote{HPO methods sometimes also consider some architectural hyperparameters but often specialize in optimizing (few) continuous hyperparameters rather than the discrete choices characteristic of NAS.} There exists little work on \emph{jointly} optimizing hyperparameter configurations and neural network architectures~\citep{Zela18,dong2020autohas,ZimLin2021a}, even though it seems natural that different architectures require different hyperparameter configurations to yield optimal performance. Indeed, there is evidence that this is actually the case. 

For example, \citet{DBLP:journals/corr/Gastaldi17} showed that the strongest version of the proposed shake-shake regularization performs best for some architectures, but is too strong for other architectures, resulting in poor performance or even divergence during training. Thus, a joint optimization of hyperparameter configurations and architecture can be expected to be beneficial.

While NAS and HPO methods typically optimize for accuracy, in many real-world applications there is more than one objective. Common objectives next to accuracy are, e.g., memory requirements, energy consumption or latency on the target hardware where the neural network is eventually deployed.

In this paper, we take a first step in the direction of \emph{multi-objective joint NAS + HPO} by proposing and empirically evaluating a set of simple, yet powerful baseline methods. All our baseline methods essentially extend current NAS or HPO approaches to cover both classical and architectural hyperparameters, optimized under multiple objectives; see Table~\ref{tbl:overview} for an overview of the methods we propose.

\begin{table}[]
\resizebox{1.0\linewidth}{!}{
\centering
\begin{tabular}{ccc}
\toprule
Proposed Method & Based on & Extended by \\ 
\midrule
    \emoash{}   (Sec. \ref{ss:emoash})    &   multi. obj. evolution \  &    successive halving    \\
     \mobohb{}    (Sec. \ref{ss:mobohb})       &   BOHB     &  multi-objective for candidate selection, MOTPE          \\
       \mdehvi{} (Sec. \ref{ss:mdehvi})         &   BO with EHVI      &   no surrogate for cheap objectives     \\
      \mobananas{} (Sec. \ref{ss:mobananas})         &  BANANAS   &  multi-objective candidate selection, successive halving    \\
     \bulkcut{}  (Sec. \ref{ss:bulkcut})         &  EA, BO        &      network morphism, pruning with knowledge distillation, constrained BO\\
\bottomrule
\end{tabular}}
\caption{Overview of the proposed methods.}
\label{tbl:overview}

\end{table}

\section{Related Work and Background}
\label{sec:rw}

\paragraph{Neural Architecture Search (NAS).} NAS refers to the task of automatically learning neural network architectures from data~\citep{elsken_survey}. NAS approaches often employ black-box optimization methods, such as evolutionary algorithms~\citep{Real17, real_regularized_2018}, reinforcement learning~\citep{zoph-iclr17a}, or Bayesian optimization~\citep{mendoza-automl16a,kandasamy_neural_2018, white2019bananas}. However, due to the large computational costs, researchers have developed methods tailored towards NAS, e.g., (gradient-based) optimization on one-shot models~\citep{bender_icml:2018,Pham18, darts}. 

\paragraph{Hyperparameter Optimization (HPO).}

It is widely acknowledged that tuned hyperparameter configurations can improve the performance of machine learning models.
Tuning manually, however, is a tedious and error-prone task.
The field of HPO~\citep[see e.g., ][for an overview]{feurer-automlbook19a} automates the search for well performing hyperparameter configurations for the data at hand.
Commonly, HPO is treated as a black-box optimization.
% Evaluating a hyperparameter configuration $\cand$ is normally considered as an expensive black box function. 
Bayesian Optimization (BO)~\citep{brochu-arXiv10a,shahriari-ieee16a} is a popular 
framework for global optimization of expensive black-box functions and has shown great success in HPO \citep[see e.g., ][]{snoek-nips12a,snoek-icml14a,feurer-aaai15a,springenberg-nips2016,eriksson-neurips19,kandasamy-jmlr20}.
BO models the expensive function using a cheap-to-evaluate probabilistic surrogate model and uses an acquisition function to trade-off exploration and exploitation for selecting a new candidate point.
% BO iteratively select a point with two key components: a surrogate model $\surro$ and an acquisition function $\acq$.
Gaussian Processes (GP)~\citep{rasmussen-book06a} and expected improvement (EI)~\citep{movckus1975bayesian} 
%\begin{equation}
%    EI_{\surro_{min}}(\conf) = \int_{-\infty}^{\infty} \max (\surro_{min} - y, 0) \prob(y|\conf)  
%\end{equation} 
are the most common choices for these.
% among different surrogate models ~\citep{hutter-lion11a, snoek-icml15a} and acquisition functions ~\citep{srninivas-icml10a, hennig-jmlr12a}.
% However, evaluating $\prob (y| \conf)$ with GP quickly becomes prohibitive as the number of observations and dimensions grow.
The tree-structured parzen estimator (TPE)~\citep{bergstra-nips11a} is an alternative surrogate that models density functions of good and bad hyperparameter configurations, respectively; optimizing the ratio of these densities is equivalent to optimizing EI~\citep{bergstra-nips11a}.
%$\surro$ fits all the previous observations and predict the possible distribution $\prob ({y|\conf})$ of the candidiates, whereas $\acq$ determines the utility of the target points $\conf$.
%Hyperopt ~\citep{bergstra-nips11a, bergstra-icml13a} used TPE to tune the hyperparameter of a neural network, while  Auto-Net ~\citep{mendoza-automl16a} automatically configures neural networks with SMAC ~\citep{hutter-lion11a}. 

\paragraph{Multi-fidelity Optimization.}
As fully training and evaluating ML pipelines (e.g., deep neural networks) can be too expensive to evaluate many configurations, multi-fidelity methods employ cheaper fidelities to reduce this cost, e.g., training only on a small subset of the data~\citep{klein-aistats17} or for a few epochs~\citep{falkner-icml18a}.
Successive Halving (SH)~\citep{jamieson2016non} and Hyperband (HB)~\citep{li-jmlr18a} are two powerful multi-fidelity strategies that allocate more budgets on the well-performing hyperparameter configurations and achieve strong anytime performance.
However, both strategies select new hyperparamter configuration at random without exploiting the knowledge gained about well-performing regions.
% the promising regions and thus only have poor final performance. 
BOHB~\citep{falkner-icml18a}, which combines BO and HB, overcomes this issue by introducing a TPE model into Hyperband to guide the search.

\paragraph{Joint NAS + HPO.}
Few researchers so far have considered the joint optimization of architectures and hyperparameter configurations. \citet{domhan-ijcai15} and \citet{mendoza-automl16a} used the random-forest-based blackbox BO method SMAC~\citep{hutter-lion11a} to jointly optimize both architectures and hyperparameter configurations (e.g., number of filters, number of layers/blocks, and conditional layer hyperparameters that are only active if a layer exists). \citet{Zela18}, \citet{runge2019learning} and \citet{ZimLin2021a} employed the more efficient multi-fidelity method BOHB~\citep{falkner-icml18a} to achieve the same goal. \citet{SMB19} first employ DARTS~\citep{darts} to search for better architectures for the task of disparity estimation and then optimize the hyperparameters of the resulting architecture with BOHB. Finally, \citet{dong2020autohas} extended NAS methods using one-shot models to also consider hyperparameters.

\paragraph{Multi-objective Optimization}
Multi-objective optimization (e.g., \citet{miettinen_evolutionary_1999}) deals with the problem of minimizing multiple objective functions $f_1(\lambda),\dots, f_n(\lambda)$. %Let $\mathcal{N}$ be the space of feasible solutions $N$ (in our case the space of feasible neural architectures). 
%In general, multi-objective optimization deals with finding $N^* \in \mathcal{N}$ that minimizes the objectives $f_1, \dots, f_n$. 
In general, there is no single $\lambda$ that minimizes all objectives at the same time since the objectives are often in competition with each other. Rather, there are multiple \emph{Pareto-optimal} solutions that are optimal in the sense that one cannot reduce any $f_i$ without increasing at least one other $f_j$ ($i \neq j$). %A solution $\lambda_1$ Pareto-dominates another solution $\lambda_2$ if $f_i(\lambda_1) \leq f_i(\lambda_2)$ for every $i$ and $f_j(\lambda_1) < f_j(\lambda_2)$ for at least one $j$. 
The set of Pareto-optimal solutions is called the \emph{Pareto front}.

\paragraph{Multi-objective Optimization with Evolutionary Algorithms.}
One class of algorithms for solving multi-objective problems are evolutionary algorithms, e.g.,~\citet{Deb2015}. Criteria for selecting candidates being mutated and defining the best current solutions are typically based on \emph{non-dominated sorting} (NDS)~\citep{6791727,deb_fast_2002} and the \emph{hypervolume indicator}~\citep{10.1007/978-3-540-31880-4_5,beume2007sms,bader2011hype}; we describe these techniques in the following. 

\emph{NDS} extends the ranking of a set of candidates 
%in a population $\mathcal{P}$
based on a single objective to multiple objectives $\objectives = (f_1,\dots,f_n)$ in the following way:
\begin{itemize}
    \item Compute the Pareto front $\mathcal{F}_1 = pareto\_front(\popul|\objectives)$ of the current population $\popul$ and assign all members of this Pareto front $\mathcal{F}_1$ the best rank.
    \item Remove the previous Pareto front from the population $\popul$ and compute the Pareto front for the remaining population: $\mathcal{F}_2 = pareto\_front(\popul \symbol{92} \mathcal{F}_1  |\objectives)$. Members of this new Pareto front $\mathcal{F}_2$ are assigned the second best rank.
    \item Iterate this process until all members of the population have been assigned a rank.
\end{itemize} 
Thus, a run of NDS partitions the population into sets $\mathcal{F}_1, \dots, \mathcal{F}_k$, where a candidate $\lambda \in \mathcal{F}_i $ outperforms another candidate  $\lambda^{\prime} \in \mathcal{F}_j $ with respect to all objectives if $i<j$.

The \emph{hypervolume indicator} $I_H$ of a population measures, informally speaking, the space of objective function values covered by the population; maximizing the hypervolume indicator corresponds to improving the Pareto front and finding better solutions. Based on the hypervolume indicator, the \emph{hypervolume subset selection problem} (HSSP)~\citep{bader2011hype} is defined as the problem of finding a subset $\popul_{HSPP} \subset \popul$ 
of a certain size $k$ 
%of a population 
so that the hypervolume is maximized for this subset: $\popul_{HSPP} \in \argmax_{ \popul^{\prime} \subset \popul, |\popul^{\prime}|= k} I_H(\popul^{\prime})$. The HSSP can also be solved to identify a poorly performing candidate by setting $k= |\popul| -1 $ and choosing the poor candidate $\lambda_{poor}$ as the one that gets removed from the population via HSSP: $\{ \lambda_{poor} \} = \popul \setminus \popul_{HSPP}$. We refer to \citet{bader2011hype} for a more formal introduction.

\paragraph{Multi-objective Optimization with Bayesian Optimization (BO).}
Traditional BO approaches only consider single objectives. However, extending BO to the multi-objective case only requires a few modifications.
For example, similar to how EI considers improvements of the objective based on a surrogate, Expected Hypervolume Improvement (EHVI)~\citep{emmerich2005single} considers contributions to the Pareto front based on a surrogate function.
EHVI has become a widely used acquisition function for multi-criteria problems due to recent works reducing the computation overhead for its calculation~\citep{yang2019multi, daulton2020differentiable}.
Recently, \citet{ozaki2020multiobjective} also extended TPE to multi-objective TPE (MOTPE) by using EHVI

and a multi-objective splitting criterium to estimate the densities used in TPE.

\paragraph{Multi-objective and constrained NAS.}
\emph{Hardware-aware} NAS has recently emerged as an important criterion for neural architectures, since many real-world applications require efficient architectures w.r.t., e.g., memory requirements, energy consumption or latency on the target hardware where the neural network is eventually deployed. Consequently, a line of research frames NAS either as a constrained~\citep{mnasnet_tan,cai2018proxylessnas} or multi-objective~\citep{Elsken19,lu2019nsganet, schorn2019automated} optimization problem to take this into account, we refer to \citet{benmeziane2021comprehensive} for a recent survey.

\section{Proposed Methods}
\label{sec:methods}

In the following, we propose five simple, yet powerful extensions of existing HPO and NAS optimization techniques to multi-objective joint HPO + NAS.

\subsection{\emoash{}: Speeding up Evolutionary Multi-Objective Algorithms}
\label{ss:emoash}
The flexibility and conceptual simplicity of evolutionary algorithms make them directly applicable to multi-objective optimization problems. For example, the well-known SMS-EMOA from~\cite{beume2007sms} is an % $(\mu + 1)$ 
evolutionary algorithm that evaluates the performance of each candidate based on its contribution to the dominated hypervolume. Although effective, evolutionary algorithms tend to be very sample-inefficient, making them too computationally expensive for many practical applications. In order to deal with this problem, we propose \emoash{} to speed up evolutionary multi-objective algorithms (EMOA) by using a multi-fidelity approach based on successive halving, see Algorithm~\ref{alg:emoash}. 

After initializing the population, we iterate EMOA with doubling the training budgets in each iteration, while the number of candidates is halved. Thus, many candidates are evaluated with a small budget to cover a wide range of solutions, while only well-performing candidates proceed to the next stage, and are evaluated with the next higher budget and used to generate new candidates. The population size remains constant, so to remove a candidate from the population we first use non-dominated sorting (NDS) to identify the subset of the population with lowest rank and then solve the hypervolume subset selection problem (HSSP) on this lowest-rank subset to  identify a poorly performing candidate, which is removed from the population.

\begin{algorithm}
\DontPrintSemicolon
\SetKwInOut{Input}{Input}
\SetKwInOut{Output}{Output}
\Input{number of function evaluations $\fe^{total}$ , population size $\pop$, maximum budget $b_{max}$, number of SH iterations $n$, objectives $\objectives$}
\Output{Pareto front w.r.t. $\objectives$}
Generate initial population $\popul$ of size $\pop$ \;
$b \leftarrow \lfloor b_{max} / 2^{n-1} \rfloor   $\tcc*[r]{initial budget}
$\fe \leftarrow \lfloor \fe^{total} / \sum_{i=0}^{n-1} 2^{-i} \rfloor   $\tcc*[r]{number of FE for the initial budget}
\For{$i=1$ \KwTo $n$}{
    Evaluate $\objectives(\lambda)$ for all $\lambda \in \popul$ with budget $b$\;
    \For{$j=1$ \KwTo $\fe - \pop$ \tcc*[r]{generate candidates for remaining FEs}}{ 
        Generate new candidate $\cand$ \tcc*[r]{parent selection and variation}
        Evaluate $\objectives(\cand)$ on budget $b$\;
        $[ \mathcal{F}_1, ..., \mathcal{F}_k ] \leftarrow \text{NDS}(\popul \cup \{ \big(\cand, \objectives(\cand) \big) \})$\; 
        $\lambda_{\text{poor}} \leftarrow \text{HSSP}(\mathcal{F}_k, |\mathcal{F}_k|-1)$\;
        $\popul \leftarrow (\popul \cup \{   \big(\cand, \objectives(\cand) \big) \}) \setminus \{ \big(\lambda_{\text{poor}}, \objectives(\lambda_{\text{poor}}) \big)  \}$\;
    }
    $ \fe \leftarrow \fe / 2$\tcc*[r]{half number of FE in next budget}
    $ b \leftarrow 2b$\tcc*[r]{double budget}
}
\Return $\pf(\popul|\objectives)$
\caption{\emoash{}\label{alg:emoash}}

\end{algorithm}
 Even though \emoash{} can be expected to require a high number of FEs, it is conceptually simple, flexible, and highly parallelizable.

\subsection{\mobohb{}: Generalization of BOHB to an Arbitrary Number of Objectives}
\label{ss:mobohb}

In order to extend BOHB~\citep{falkner-icml18a} to multi-objective optimization, we apply two changes, one related to the BO part and one related to the HB part. Firstly, we replace the TPE~\citep{bergstra-nips11a} originally used in BOHB by MOTPE~\citep{ozaki2020multiobjective} to guide the search by selecting new configurations under consideration of multiple objectives.
Secondly, we extend HB~\citep{li-jmlr18a} in a similar fashion as for \emoash{} and MOTPE to decide with which configuration to proceed in the next stage: we use NDS and the result of the HSSP~\citep{bader2011hype}.

Our proposed \mobohb{} generalizes BOHB to an arbitrary number of objectives by making the aforementioned modifications; in particular, a single objective is only a special case. Note that this approach can also be used to extend to other methods combining a (multi-objective) black-box optimizer and multi-fidelity optimization strategy.

\subsection{\mdehvi{}: Mixed Surrogate Expected Hypervolume Improvement}
\label{ss:mdehvi}

Although EHVI~\citep{emmerich2005single} can be directly applied for joint NAS and HPO obtaining competent results, we further enhance the algorithm by a simple observation from \citet{Elsken19}: while some of the objective functions are expensive to evaluate (e.g., evaluating the accuracy is expensive since it requires training the network first), other are cheap to evaluate (e.g., the number of parameters). Thus, rather than relying on a surrogate model for \emph{every} objective function as in vanilla EHVI, we solely use surrogate models for the expensive objective function and directly evaluate the cheap objectives. This way, we avoid fitting surrogate models for objectives which are cheap to evaluate anyway and avoid poor predictions of the surrogate models.

We refer to Algorithm~\ref{alg:mdehvi} for pseudo-code. The approach is very similar to performing BO with EHVI, but, instead of fitting surrogate models for each considered dimension, we provide true evaluations for the cheap objectives.

\begin{algorithm}
\DontPrintSemicolon
\SetKwInOut{Input}{Input}
\SetKwInOut{Output}{Output}
\Input{expensive and cheap objectives $\objectives = (\fexp, \fcheap)$, surrogate model $\surr$, number of function evaluations $\fe$}
\Output{Pareto front w.r.t. $\objectives$}
Initialize population $\popul$ with initial observations \;
\For{$k=1$ \KwTo $\fe$}{
    Fit surrogate model $\surr$ on $\popul$ \;
    Select next candidate: $\cand \in arg max_{\lambda}\text{EHVI}\big(\lambda| \popul, \surr, \fcheap \big) $ \;
    Evaluate $ \fexp(\cand)$ \;
    Update data: $\popul \leftarrow \popul \cup \Big\{ \big( \cand, \objectives(\cand) \big) \Big\}$\;
}
\Return $\pf(\popul|\objectives)$
\caption{\mdehvi{}\label{alg:mdehvi}}
\end{algorithm}

Empirical results show how our method is able to obtain an appropriate exploration of the Pareto front, effectively showing a superior performance compared to vanilla EHVI, compare Figure \ref{fig:ehvi_basline} in the appendix.

\subsection{\mobananas{}}
\label{ss:mobananas}

BANANAS~\citep{white2019bananas} uses an ensemble of neural networks for predicting the performance of a neural network within BO in combination with a novel path-based encoding. 
Given a set of already evaluated candidates, the best one is chosen and mutated to generate new candidates. These candidates are evaluated by means of the predictor, and the resulting predictions are used to determine the next candidate for evaluation in combination with an acquisition function as usual in BO.

To extend BANANAS for multi-objective optimization, we create a Pareto front from the predictions for the objective function values and sort according to crowd sorting~\citep{RN72005}. Subsequently, the next architectures are drawn from this ranking by means of independent Thompson sampling for evaluation, which works best for candidate selection according to \citet{white2019bananas}. In the same way, parent architectures are selected for generating new candidates through mutations by first forming a Pareto front and then sorting it using crowd sorting, see Algorithm \ref{alg:MOBANANAS}.

Note that we do not use the path-based encoding from \citet{white2019bananas} since it is not meaningful for our search space. Rather, we employ a simple real-valued vector representation, which furthermore also directly allows us to include non-architectural hyperparameters. We employ Gaussian noise for mutating parents: 
We assume integer-valued hyperparameters (e.g., number of layers), and normalize each value by dividing by the maximum value to map each hyperparameter to the range $[0,1]$: we then add Gaussian noise.
Before each function evaluation, this continuous representation is discretized by choosing the integer-valued hyperparameter which is closed to the mutation value after normalization.

\begin{algorithm}
\DontPrintSemicolon
\SetKwInOut{Input}{Input}
\SetKwInOut{Output}{Output}
\Input{neural predictor $\neuralpred$, number of candidates to mutate $n_{\text{mut}}$, mutation variance $\sigma^2$, number of new candidates $n_{\text{new}}$, objectives $\objectives$}
\Output{Pareto front w.r.t. $\objectives$}
Generate initial population $\popul$\;
\For{$i=1$ \KwTo $n$}{
train neural predictor $\neuralpred$ on $\popul$\;
sort $\popul$ using  $\text{NDS}(\popul)$ and crowdingDistance$(\popul)$\;
choose top-$n_{\text{mut}}$ candidates from $\popul$ and \textbf{mutate} by adding noise $\eta \sim \mathcal{N}(0,\sigma^2)$ (drawn independently for each dimension of the candidates)\; 
evaluate chosen $n_{\text{mut}}$ candidates using $\neuralpred$ \;
choose top-$n_{\text{new}}$ candidates $\lambda_1, \dots, \lambda_{n_{\text{new}}}$ via independent Thompson sampling\;
evaluate  $\objectives(\lambda_1), \dots, \objectives(\lambda_{n_{\text{new}}})$
$\popul  \leftarrow \popul \cup \{{ \big(\lambda_1, \objectives(\lambda_1) \big), \cdots, \big(  \lambda_{n_{new}}, \objectives(\lambda_{n_{new}}) \big)       }\}  $\;
}
\Return $\pf(\popul|\objectives)$
\caption{\mobananas{}  \label{alg:MOBANANAS}}
\end{algorithm}

This approach can be further extended by successive halving to quickly discard poorly performing architectures chosen by the neural predictor. In the experimental section we employ this version of \mobananas{}  since we empirically found it to yield improved performance.

\subsection{\bulkcut{}}
\label{ss:bulkcut}

\bulkcut{} combines a very simple evolutionary strategy with BO. The name \bulkcut{} comes from the fact that the algorithm first looks for high accuracy models by successively enlarging them with network morphisms~\citep{chen2015net2net}, then shrinking them using pruning techniques in combination with knowledge distillation~\citep{hinton2015distilling}. 
A \bulkcut{} run comprises three sequential phases:%. The most important difference between them is the method used to decide the architecture of new individuals:

\begin{enumerate}
\item Initialization: sample random architectures and train them;

\item Bulk-up: generate offsprings by applying network morphisms;
\item Cut-down: prune bulked-up models.
\end{enumerate}

After the initialization phase is completed, parents are selected for the bulk-up phase. For this, we propose the \textit{Paretsilon Greedy} criterion, which 
combines non-dominated sorting and an $\epsilon$-greedy exploration strategy, as described in Algorithm \ref{alg:paretsilon}. This criterion attributes a non-zero chance for being a parent to all models from the initialization phase. However, the chance of selection is higher for individuals in fronts closer to the Pareto Front. Individuals from the same front are selected with equal probability.

\begin{algorithm}
%\DontPrintSemicolon
\SetKwInOut{Input}{Input}
\SetKwInOut{Output}{Output}
\Input{Population $\popul$, exploration probability $\epsilon$}
\Output{$\lambda \in \popul$}

\While{True}
{
	$\mathcal{F}  \gets pareto\_front(\popul)$\;
	\uIf{$rand() \leq 1-\epsilon$ or $\mathcal{F}  == \popul$ }{
	    Sample $\lambda$ from $\mathcal{F}$\;
    	\Return $\lambda$\;
  	}
%   	\uElseIf{$rand() \leq 1-\epsilon$}{
%   		Sample $\lambda$ from $\mathcal{F}$ \;
%     	\Return $\lambda$\;}
  	\Else{
      	$\popul \gets \popul \setminus \mathcal{F} $;
  	}
}
\caption{Paretsilon greedy}
\label{alg:paretsilon}
\end{algorithm}

Once a parent is chosen, an offspring is generated by applying a network morphism. Network morphisms are commonly employed as mutations in the NAS literature since they avoid retraining from scratch by inheriting the knowledge of the parent~\citep{Elsken17,Elsken19, schorn2019automated,cai2017reinforcement}. In our experiments, two morphism operators are implemented: insert a convolutional layer and insert a fully-connected layer, at random positions.

In the cut-down phase, we employ pruning techniques to shrink models from the first two phases. 
We use structured pruning~\citep{Anwar_structured} in our experiments, i.e., eliminating units from fully-connected layers and filters from convolutional layers, instead of dropping individual weights.

The units/filters were ranked by the sum of their output weights, and those on the bottom of the list were pruned.

We employ knowledge distillation~\citep{hinton2015distilling} for training the shrunken models so that they match their parent's output~\citep{Elsken19,Prakosa2020ImprovingTA,chen_kd}. Note that in the evaluation phase of these models, there is no more training and the shrunk models are not fine-tuned on training labels.

During all three phases, the non-architectural hyperparameters (e.g., learning rate and weight decay) are optimized via constrained BO. The term \textit{constrained} here refers to the fact that the optimization of the acquisition function is performed with constrains on all architectural hyperparameters, since they have already been specified, as previously explained. Note that the surrogate model still covers the architectural hyperparameters; thus it is still aware of them and the surrogate model is fitted and shared across architectures.

Algorithm \ref{alg:bnc} summarizes how \bulkcut{} works.

\begin{algorithm}
\DontPrintSemicolon
\SetKwInOut{Input}{Input}
\SetKwInOut{Output}{Output}
\Input{time budgets $T_1 < T_2< T_3$, exploration probability $\epsilon$, objectives $\objectives$}
\Output{set of Pareto optimal solutions}
$\popul \gets \emptyset$      //population set\;
$t \gets elapsed\_time()$ \;
\While{$t <T_3$}{
 \uIf{ $ t \in [ 0, T_1 ) $}{
  $\lambda_{\alpha} \gets random\_architecture()$\;
}
 \uIf{ $ t \in [ T_1, T_2 ) $}{
    $\lambda_{\alpha} \gets paretsilon\_greedy(\popul,\epsilon)$ \tcc*[r]{parent selection (see Alg.~\ref{alg:paretsilon})}
      $\lambda_{\alpha} \gets network\_morphism(\lambda_{\alpha})$\;
 }
 \uIf{ $ t \in [ T_2, T_3 ] $}{
      $\lambda_{\alpha} \gets paretsilon\_greedy(\popul,\epsilon)$ \tcc*[r]{parent selection (see Alg.~\ref{alg:paretsilon})}
    $\lambda_{\alpha}  \gets prune\_and\_distill\_knowledge(\lambda_{\alpha})$\;
}
  $\lambda_{\beta} \gets constrained\_BO(\lambda_{\alpha} )$ \tcc*[r]{other hyperparameters}
  evaluate $\objectives(\lambda_{\alpha}, \lambda_{\beta}) $	\;  
  Update $constrained\_BO$ with $\big(\lambda_{\alpha}, \lambda_{\beta}, \objectives(\lambda_{\alpha}, \lambda_{\beta})\big)$ \;  
  $\popul \gets \popul \cup \{\big(\lambda_{\alpha}, \lambda_{\beta}, \objectives(\lambda_{\alpha}, \lambda_{\beta})\big) \}$\;
  $t \gets elapsed\_time()$ \;
}
\Return $pareto\_front(\popul)$
\caption{\bulkcut{}}
\label{alg:bnc}
\end{algorithm}

\section{Experiments}
\label{sec:experiments}

We start by describing our experimental setup in Section \ref{sec:esetup} and present results in Section \ref{sec:results}.

\subsection{Experimental Setup}
\label{sec:esetup}
Each of the proposed methods, along with random search as the simplest baseline, is run 10 times on each dataset, with a runtime limit of 24 hours (per run and per method) on a single RTX 2080 Ti GPU.

A maximum budget of 25 epochs for training a single configuration is defined, although it is up to the methods to decide if they want to train with a smaller budget to speed up the search. 
We target network size (by means of number of parameters) and classification accuracy as the objectives of our multi-objective optimization.

\paragraph{Datasets.}
 We used the Oxford-Flowers dataset~\citep{Nilsback06}, a small dataset composed of 17 different classes with 80 examples each, to show the performance of the proposed approaches in environments where many, cheap function evaluations are available. All images are scaled down to a resolution of $16x16$ for computational reasons. We also tested our methods on Fashion MNIST~\citep{xiao2017/online}. 
We split the datasets as follows: for Flowers, we randomly split the data into 60\% for training, 20\% for validation and 20\% for testing. For Fashion-MNIST, we split the train set as defined in PyTorch into training (80\%) and validation (20\%) and use the original test split for testing. Neural network weights are always trained on training data, their performance on validation data is used to guide hyperparameter and architecture optimization, and the test set is only used for evaluation.

\paragraph{Search Space.}
As both target datasets are composed of images as input, all architectures start with a variable number of convolutional layers, with a variable number of filters for each layer and a variable kernel size. All layers employ ReLU activation functions, followed by max-pooling to reduce spatial dimensions. Batch normalization may be chosen after each layer. After the last convolution, a global average pooling may be applied. The feature volume is flattened and fed to a variable-length sequence of fully connected layers, with a variable number of neurons and using ReLU as the activation function. After the last hidden layer, a final fully connected layers maps to the class predictions.
We always use Adam~\citep{Kingma-iclr15} to optimize neural network weights, with %default betas of $0.9$ and $0.999$ and fixed 
a searchable learning rate % (between $10^{-5}$ and $1$) 
and batch size. The number of filters of the convolutional layers, the number of neurons on the hidden layers, the learning rate, and the batch size are considered on a logarithmic scale.

We refer to Table \ref{tbl:search_space} for a summary of our search space.

\begin{table}[]
\centering
\begin{tabular}{|l|r|r|}
\hline
Hyperparameter                         & \multicolumn{1}{l|}{Range} & \multicolumn{1}{l|}{Log scale} \\ \hline \hline
Num. convolutional layers    & $\{1,2,3\}$                      & No                             \\ \hline
Num. filters conv. layer $i$ & $[2^4, 2^{10}]$                  & Yes                            \\ \hline
Kernel size                  & $\{3, 5, 7\}$                  & No                             \\ \hline
Batch normalization          & $\{true, false\}$                     & No                             \\ \hline
Global average pooling       & $\{true, false\}$                     & No                             \\ \hline
Num. fully connected layers  &  $\{1,2,3\}$                        & No                             \\ \hline
Num. neurons FC layer $i$    & $[2^1, 2^9 ]$                    & Yes                            \\ \hline
Learning rate                & $ [10^{-5},10^{0} ]$           & Yes                            \\ \hline
Batch size                   & $[2^0, 2^9]$                    & Yes                            \\ \hline
\end{tabular}
\caption{Joint space of architectural and non-architectural hyperparameters being optimized.}
\label{tbl:search_space}

\end{table}

\begin{figure}[h]
  \centering
  \begin{subfigure}{\linewidth}
    \centering
    \includegraphics[width=.95\linewidth]{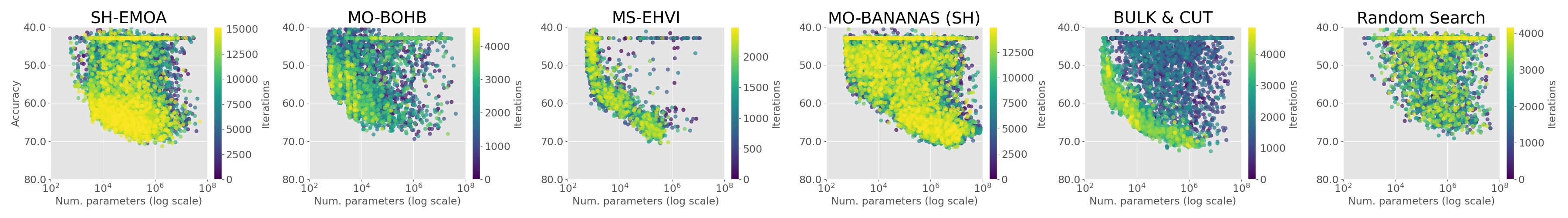}
    \caption{Sampled configurations for each method on Flowers. We only show a random subset of $10\%$ of all sampled points for visualization purposes. 
    }
  \end{subfigure}

  \begin{subfigure}{\linewidth}
    \centering
    \includegraphics[width=.95\linewidth]{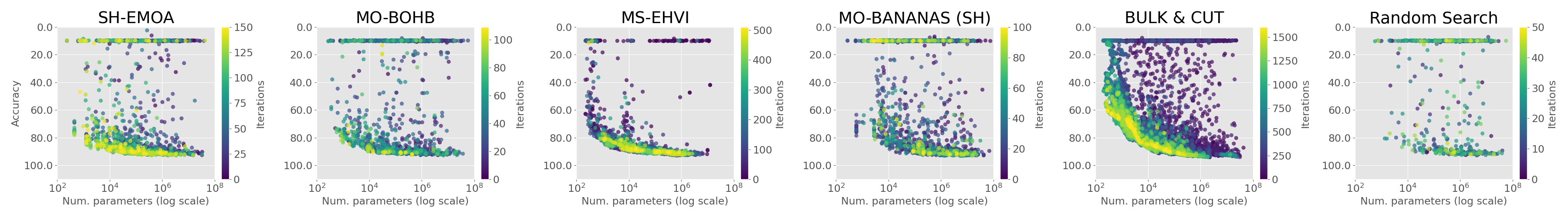}
    \caption{Sampled configurations on Fashion-MNIST dataset.}
  \end{subfigure}  
  \caption{Sampled configurations for each method on Fashion-MNIST.}
 \label{fig:sampled_configs}
\end{figure}

\subsection{Results}
\label{sec:results}

\paragraph{Visualizing sampled configurations.}
Figure \ref{fig:sampled_configs} visualizes the sampled configurations for each method across all 10 random seeds. 

For the Flowers dataset (upper row), all methods (except random search) explore the Pareto front, which is what they were designed for. However, they also significantly differ in the exploration strategy. \emoash{} explores both objective functions equally well but still samples poor configurations in later iterations. \mobohb{} tends to focus on smaller networks in later iterations, while \mdehvi{} very quickly discovers networks close to the Pareto front and mostly samples new configures there. After an initial phase, \bulkcut{} also mostly samples candidates close to the Pareto front in later iterations.
%, which is reasonable simply due to the design of the method. 
For Fashion-MNIST, already very small networks yield a high accuracy, making it hard to actually trade-off the different objective functions. However, one can still see that all methods focus on promising regions in objective function space.

\begin{figure}
  \centering
  \begin{subfigure}{0.47\linewidth}
    \centering
    \includegraphics[width=.95\linewidth]{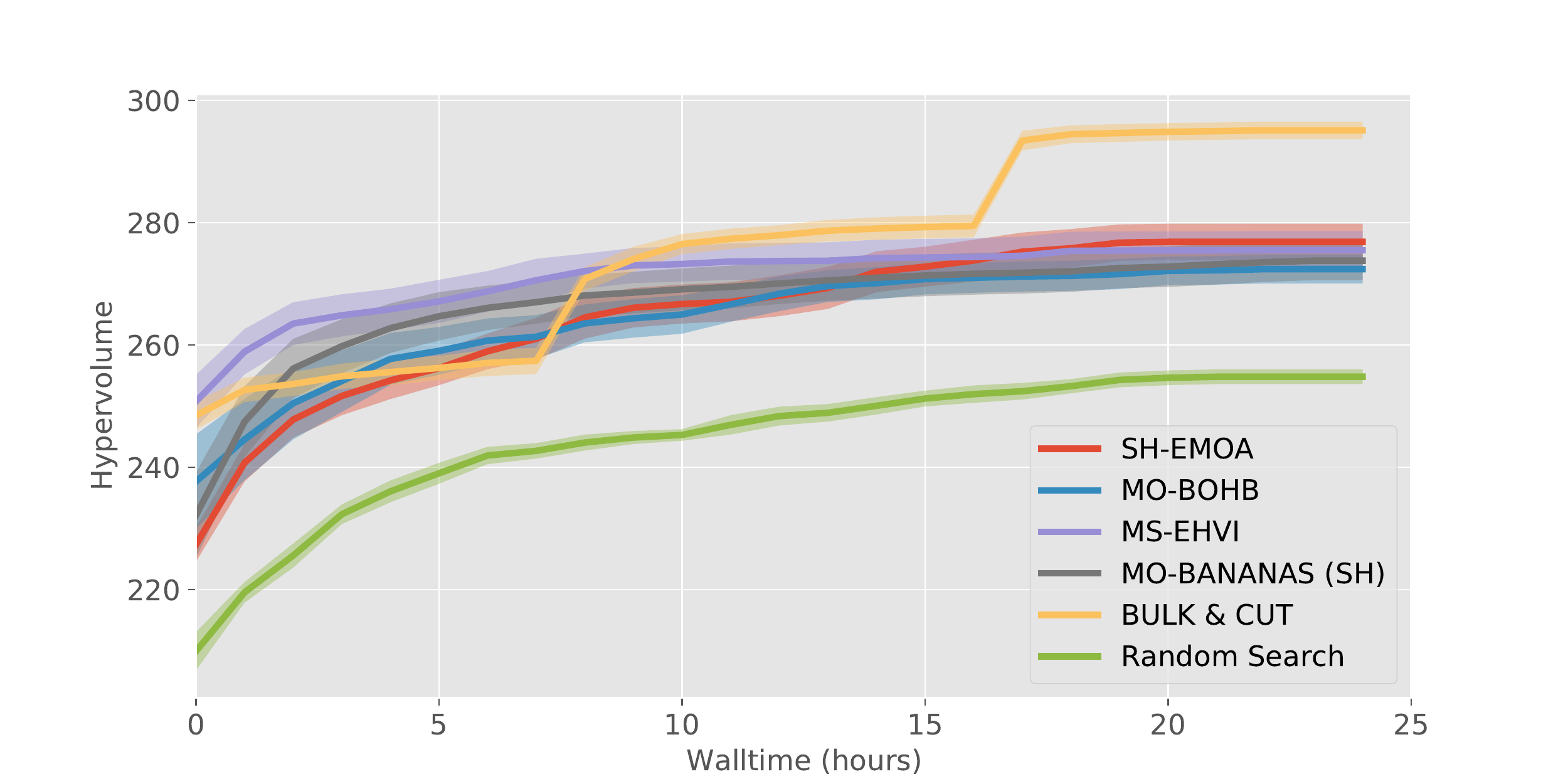}
    \caption{Flowers dataset on validation split.}
  \end{subfigure}%
  \begin{subfigure}{0.47\linewidth}
    \centering
    \includegraphics[width=.95\linewidth]{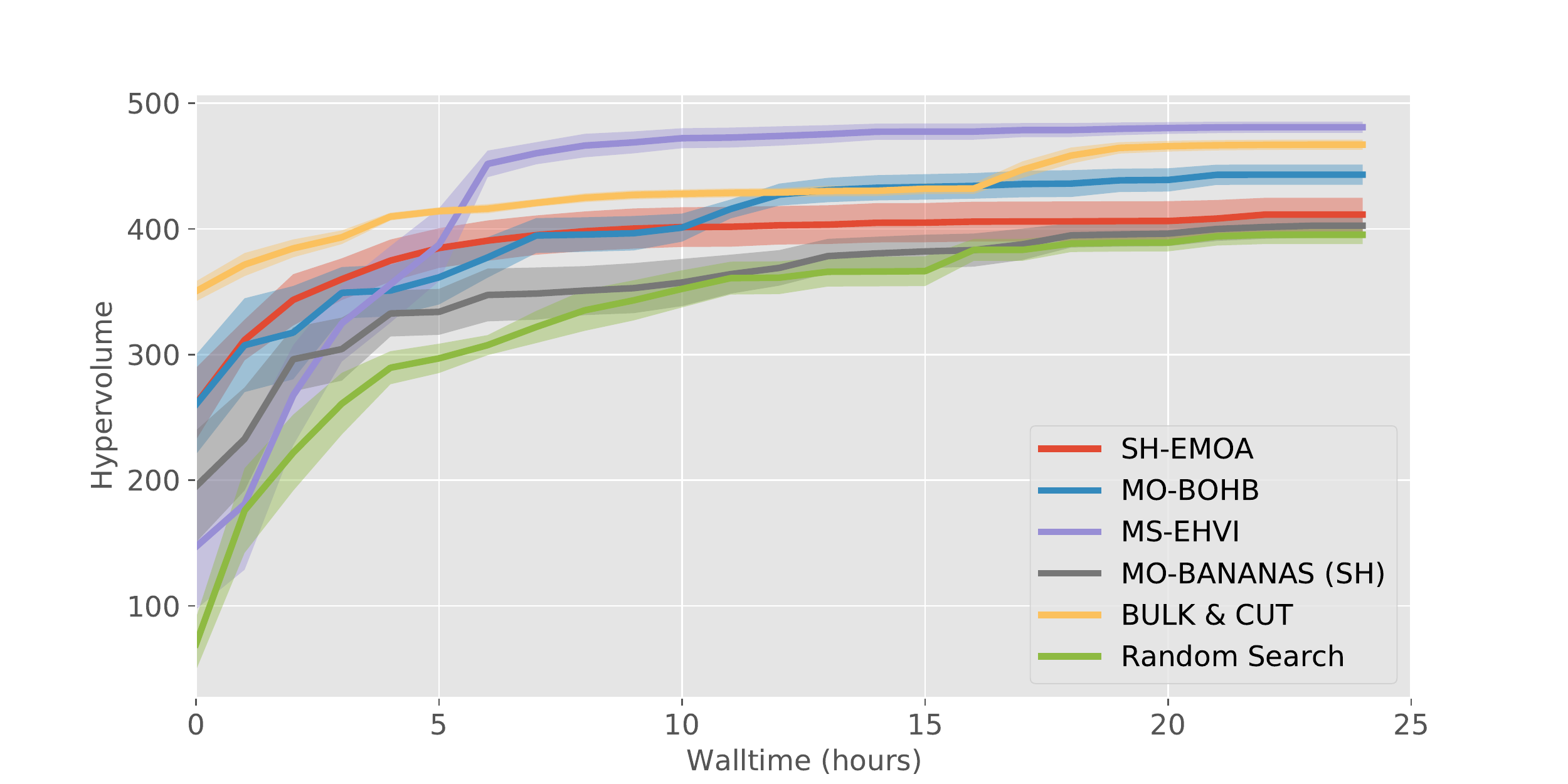}
    \caption{Fashion-MNIST dataset on validation split.}
  \end{subfigure}
  \begin{subfigure}{0.47\linewidth}
    \centering
    \includegraphics[width=.95\linewidth]{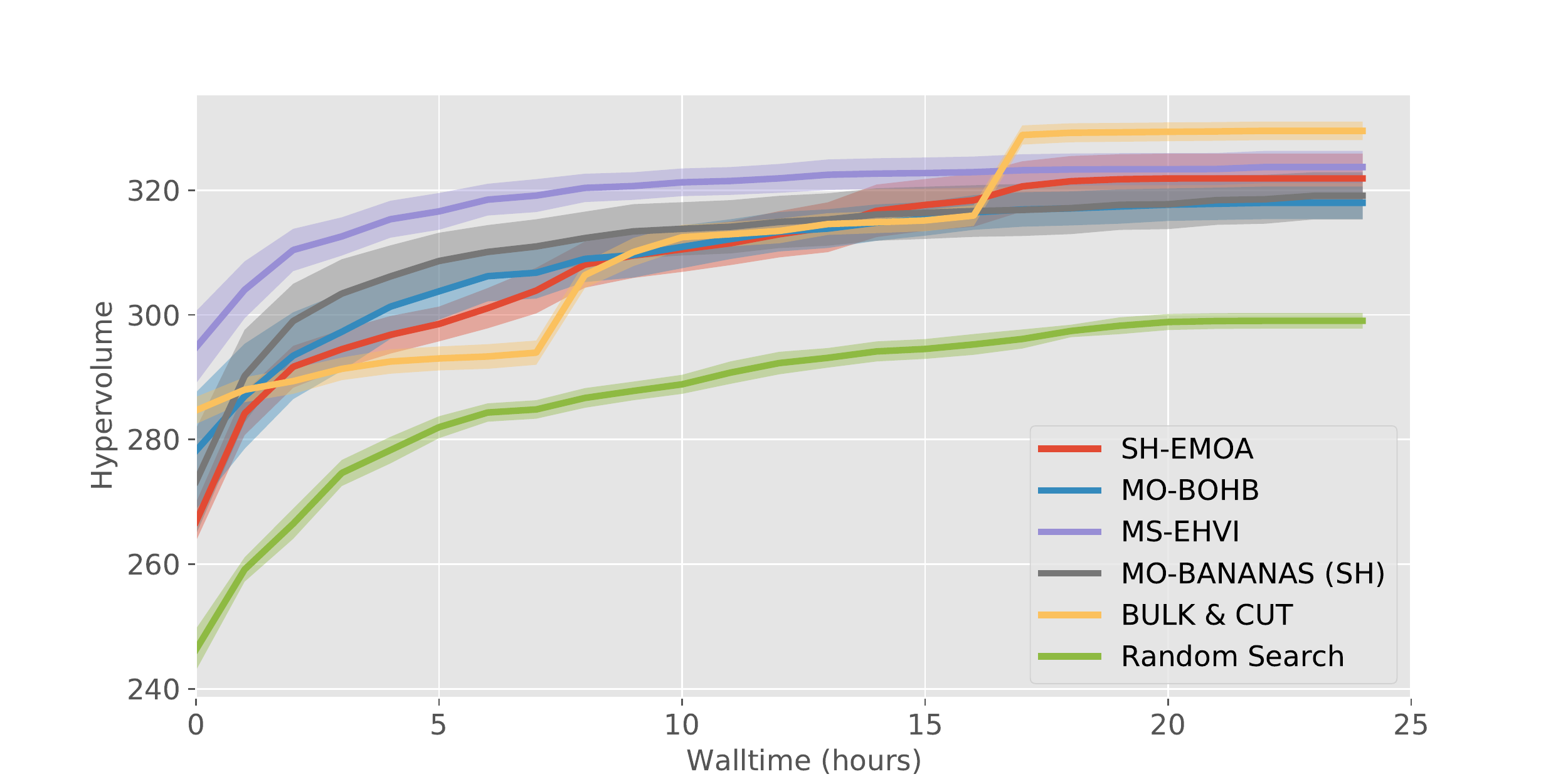}
    \caption{Flowers dataset on test split.}
  \end{subfigure}%
  \begin{subfigure}{0.47\linewidth}
    \centering
    \includegraphics[width=.95\linewidth]{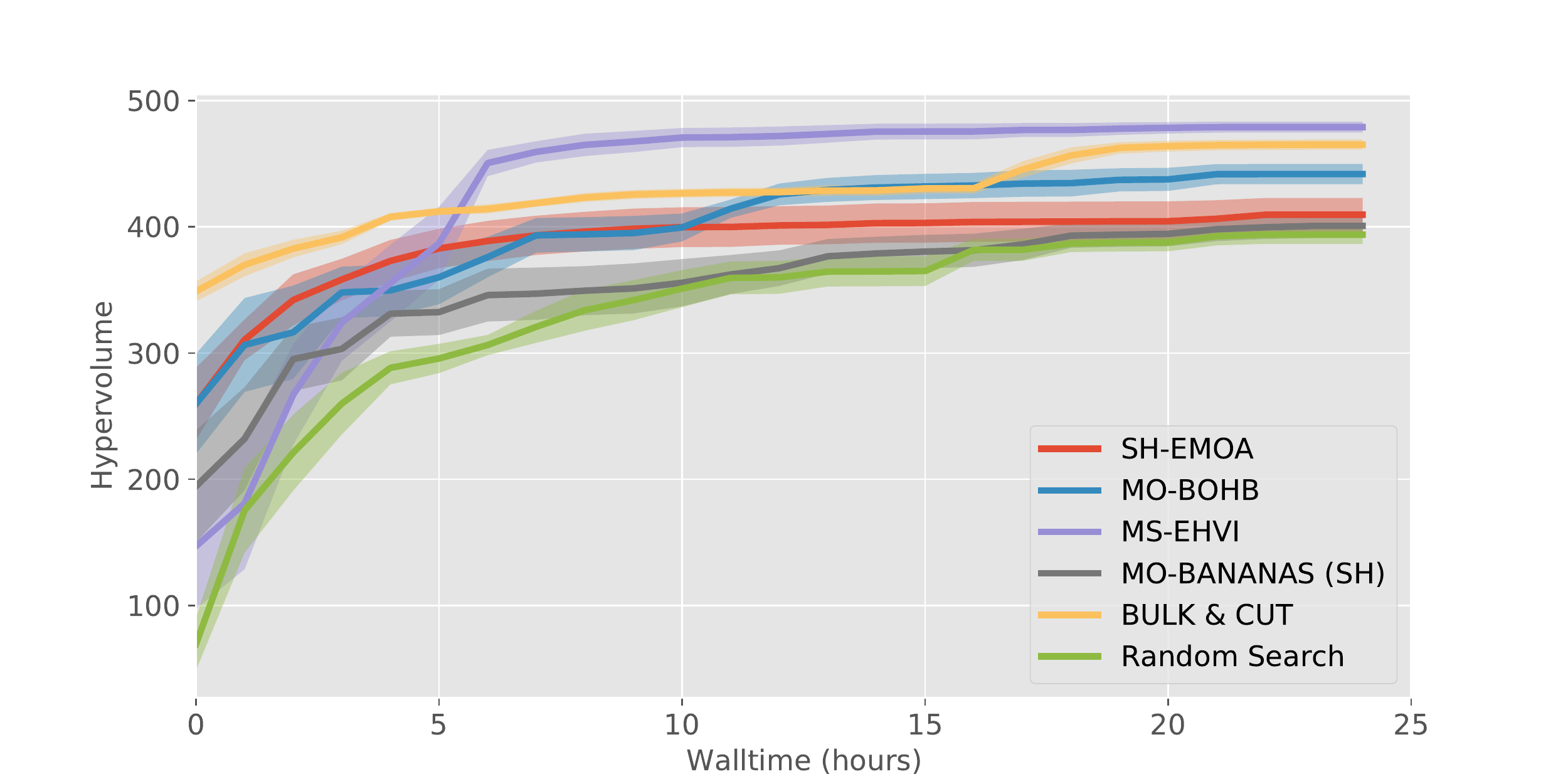}
    \caption{Fashion-MNIST dataset on test split.}
  \end{subfigure}  
  \caption{Hypervolume obtained by the different methods over time. We show means $\pm$ standard errors of the mean based on 10 independent runs of each method.
  }

\label{fig:hv}
\end{figure}

\paragraph{Anytime performance over the course of multi-objective joint NAS+HPO.}
Figure \ref{fig:hv} shows the hypervolume indicator over time, averaged across the 10 independent runs per method on each of the two datasets. All our methods clearly outperform random search. 
For Flowers, \mdehvi{} converges very fast but is eventually outperformed by \bulkcut{}, which however performs less strongly in the initial phase. \emoash, \mobohb{} and \mobananas{} perform similarly. For Fashion-MNIST, \mdehvi{} and \bulkcut{} again slightly outperform the remaining approaches.

\begin{figure}
  \centering
  \begin{subfigure}{0.45\linewidth}
    \centering
    \includegraphics[width=.95\linewidth]{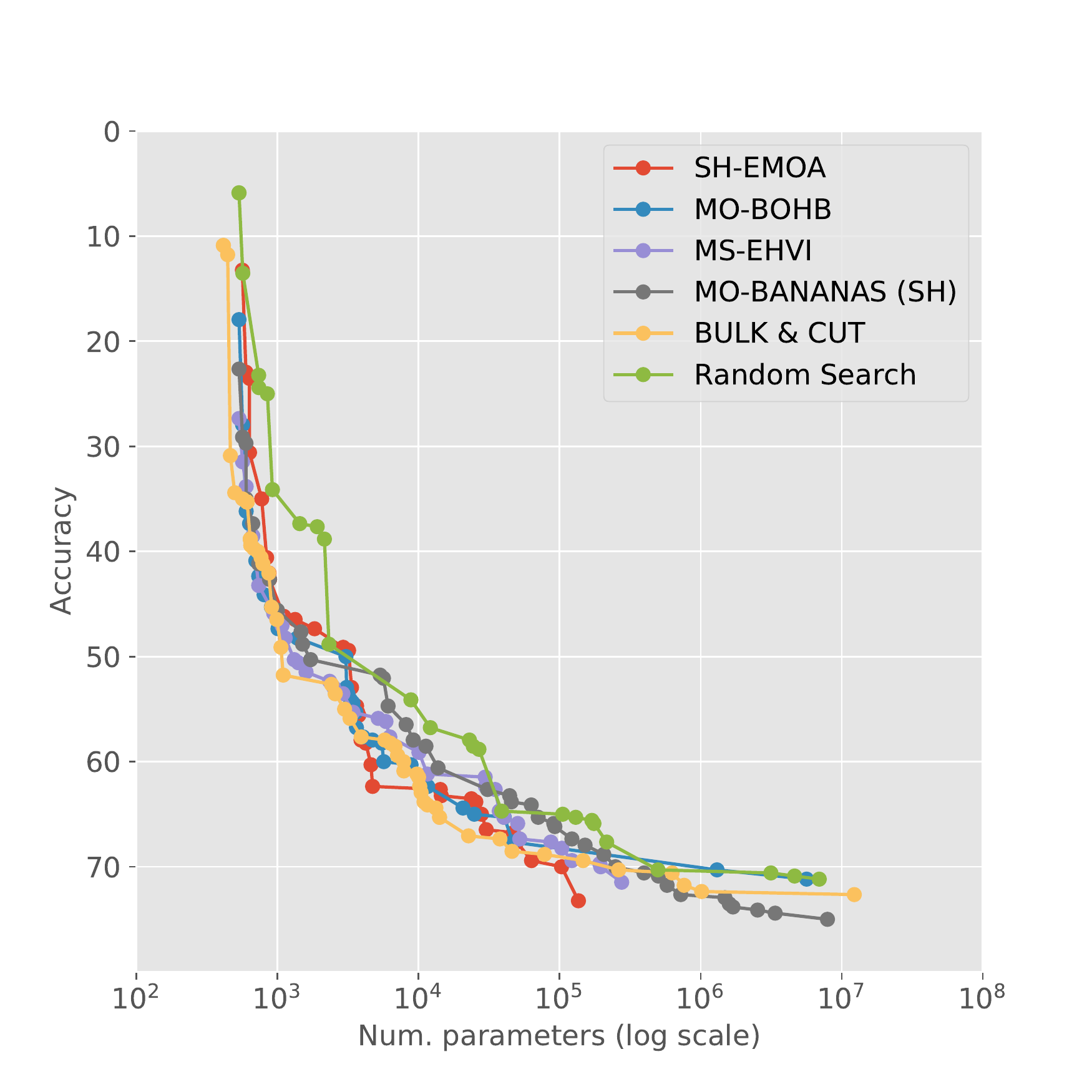}
    \caption{Pareto fronts on Flowers dataset.}
  \end{subfigure}%
  \begin{subfigure}{0.45\linewidth}
    \centering
    \includegraphics[width=.95\linewidth]{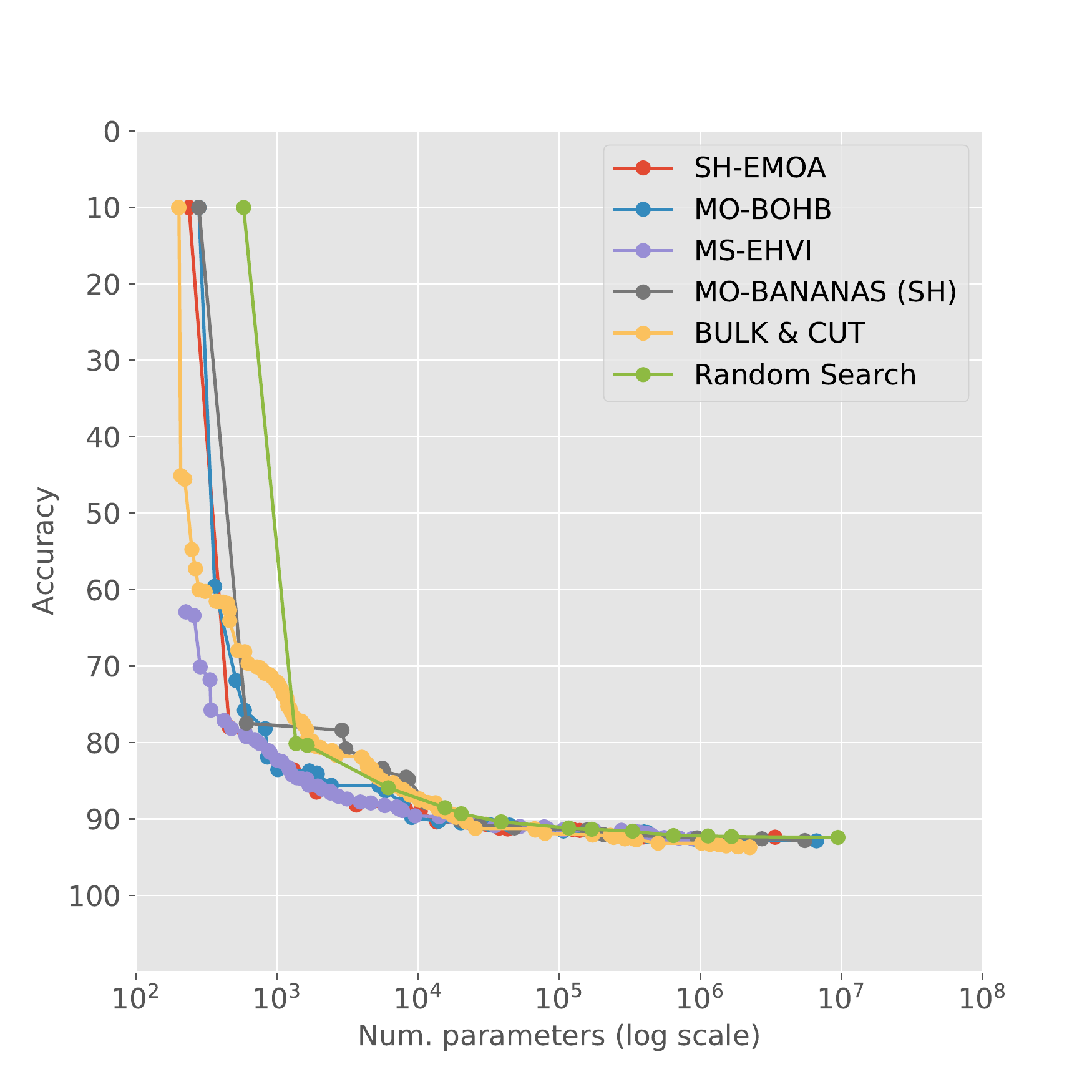}
    \caption{Pareto fronts on Fashion-MNIST dataset.}
  \end{subfigure}  
  \caption{Pareto fronts by combining the results of the 10 different runs for each method on the test split}.% Legend includes 

\label{fig:pf_final}
\end{figure}

\paragraph{Final results.}
Figure \ref{fig:pf_final} shows the final Pareto fronts for all methods when combining the results from all seeds.

The proposed methods perform similarly for the range of parameters from $10^3$ to $10^5$, but some methods have problems with covering smaller or larger models. When looking at results for each seed  other runs (please refer to Figures \ref{fig:pf_final_app_1} and \ref{fig:pf_final_app_2} in the appendix), we however also noticed that the results vary across seeds, indicating that initializing might have a high impact on the performance and that the budget of 24 hours might not be sufficient for the methods to converge or that methods simply get stuck in a local optimum.
Table \ref{tbl:test_results} summarizes the obtained hypervolume of the final Pareto front for each method.

\begin{table}[]
\centering
\begin{tabular}{l|r|r|}
\cline{2-3}
& \multicolumn{2}{c|}{Hypervolume (mean $\pm$ std. error)}                       \\ \cline{2-3} 
& \multicolumn{1}{c|}{Flowers} & \multicolumn{1}{c|}{Fashion-MNIST} \\ \hline 
\multicolumn{1}{|l|}{Random search} & 299.05 $\pm$ 1.19               & 393.86 $\pm$ 6.97                     \\ \hline
\multicolumn{1}{|l|}{\emoash{}}       & 321.90 $\pm$ 3.79               & 409.61 $\pm$ 12.61                    \\ \hline
\multicolumn{1}{|l|}{\mobohb{}}       & 317.98 $\pm$ 2.48               & 441.86 $\pm$ 7.58                     \\ \hline
\multicolumn{1}{|l|}{\mdehvi{}}       & 323.72 $\pm$ 2.46               & 479.13 $\pm$ 4.15                     \\ \hline
\multicolumn{1}{|l|}{\mobananas{}}    & 319.11 $\pm$ 3.59               & 400.71 $\pm$ 9.00                     \\ \hline
\multicolumn{1}{|l|}{\bulkcut{}}   & 329.54 $\pm$ 1.41               & 465.19 $\pm$ 3.83                     \\ \hline
\end{tabular}
\caption{Final hypervolume obtained by each method on both test datasets. We show means $\pm$ standard errors based on 10 independent runs.}
\label{tbl:test_results}
\end{table}

\section{Conclusions}
\label{sec:conc}

We addressed the problem of joint hyperparameter optimization and neural architecture search under multiple objectives by extending existing methods to this scenario. We recommend that the proposed methods serve as baselines for future research in this direction. To facilitate this, all our code is available under \url{https://github.com/automl/multi-obj-baselines} under a permissive Open Source license (Apache 2.0).

\acks{}
The authors acknowledge funding by the Robert Bosch GmbH. A part of this work was supported by the German Federal Ministry of Education and Research (BMBF, grant RenormalizedFlows 01IS19077C).

\newpage

%\bibliography{bib/strings,bib/local,bib/lib,bib/proc}
\bibliography{}

% Appendix goes to a new page

\newpage

\appendix
\section{Full Details on the Various Baselines}

\subsection{Implementation details on \emoash{}}\label{app1}
\begin{itemize}
\itemsep0em
    \item Population size and total number of samples: In Flowers dataset, we use $\pop=100$ and $\fe^{total}=15000$. In Fashion-MNIST, we use $\pop=10$ and $\fe^{total}=150$.
    \item Parent selection: We use tournament selection by randomly sampling $k$ potential parents from the current population (uniform distribution) and choose the parent with highest fitness. We use $k=3$.
    \item Variation: On each step we choose either mutation or recombination strategy with equal probability. Mutation is defined as a uniformly distributed random variation of $5$ hyperparameters from the parent configuration. For recombination, we use two parents and choose each hyperparameter from one of them with equal probability. As we have a conditional search space, relationships between hyperparameters are taken into account when a new individual is created. For example, if a mutation increases the total number of convolutional layers, the size of the kernel for each new layer is also added.
\end{itemize}

\subsection{Implementation details on \mobohb{}}\label{app:MOBOHB}
Algorithms~\ref{alg:mobohb} and~\ref{alg:mobohbSampling} show pseudo code for \mobohb{} and its sampling step, respectively. Note the close resemblance to the original BOHB (differences marked in red), as \mobohb{} generalizes BOHB to any number of objectives.
Note however, that there are two minor differences between the current version of our proposed \mobohb{} and the original BOHB implementation: (i) \mobohb{} uses an hierarchy of one-dimensional KDEs, whereas BOHB use a single multi-dimensional KDE, and (ii) we do not multiply bandwidths by a constant factor $b_w$. In future versions of \mobohb{}, we suspect that using a single multi-dimensional KDE and multiplication of bandwidths may further improve performance by better handling interaction effects between (architectural) hyperparameters and encouraging more exploration around promising configurations, respectively.
\\
\begin{minipage}[t]{0.485\textwidth}
\begin{algorithm}[H]
\DontPrintSemicolon
\SetKwInOut{Input}{Input}
\SetKwInOut{Output}{Output}
\Input{budgets $b_{min}$ and $b_{max}$, configurations discarding factor $\eta\in\mathbb{N}_{> 0}$, and objectives $\objectives$}
\Output{Pareto front w.r.t. \objectives}
$s_{max} \leftarrow \lfloor \log_{\eta} \frac{b_{max}}{b_{min}} \rfloor$\;
$\popul_{b} \leftarrow [~]\forall b\in\lbrace \eta^{-s}\cdot b_{max}| s=s_{max}, s_{max-1},...,0\rbrace$\;
\While{not stopping criterion}{
\For{$s\in\lbrace s_{max}, s_{max-1},...,0\rbrace$}{
sample $n=\lceil \frac{s_{max} + 1}{s+1}\eta^{s} \rceil$ configurations $\lambda_1,...,\lambda_n$ using Algorithm~\ref{alg:mobohbSampling}\;
run \textcolor{red}{modified SH} on $\lambda_1,...,\lambda_n$ with initial budget $\eta^{-s}\cdot b_{max}$\;
add observations $\lbrace(\lambda_i, \objectives(\lambda_i))\rbrace$ of each budget $b$ to $\popul_b$\;
}
}
\Return \textcolor{red}{$\pf(\popul_{b_{max}}|\objectives)$}
\caption{\mobohb{}\label{alg:mobohb}}
\end{algorithm}
\end{minipage}
\hspace{0.02\textwidth}
\begin{minipage}[t]{0.495\textwidth}
\begin{algorithm}[H]
\DontPrintSemicolon
\SetKwInOut{Input}{Input}
\SetKwInOut{Output}{Output}
\Input{observations $\popul$, fraction of random runs $\rho$, quantile $\gamma$, number of samples $n$, and minimum number of points $N_{min}$ to build a model}
\Output{next configuration to evaluate}
\uIf{$rand()<\rho$}{
    \Return random configuration \;
  }
 $b \leftarrow \argmax \lbrace \popul_{b}:|\popul_{b}|\geq N_{min}+2\rbrace$\;
\uIf{$b=\emptyset$}{
    \Return random configuration \;
  }
\textcolor{red}{greedily split $\popul$ into good $\popul_l$ or bad $\popul_g$ observations using NDS \& HSSP}\;
fit KDEs $l$ and $g$ based on $\popul_l$ or $\popul_g$, respectively\;
draw $n$ samples according to $l(\lambda)$\;
\Return sample with highest ratio $\frac{l(\lambda)}{g(\lambda)}$\;
\caption{Sampling in \mobohb{}\label{alg:mobohbSampling}}
\end{algorithm}
\end{minipage}

In all our experiments, we set the meta-parameters of \mobohb{} as follows: For the HB part of \mobohb{}, we set the configuration discarding factor of to $\eta=3$, use minimum budget $b_{min}= 5$ and maximum budget $b_{max}= 25$. In the BO part of \mobohb{}, we use an random fraction $\rho=1/6$, set the quantile to $\gamma=0.1$, sample $n=24$ configurations, and use a minimum of $N_{min}= 2\cdot |HPs| + 1$ points before building a model.
\subsection{Details on \mdehvi{}}

\begin{figure}
  \centering

  \begin{subfigure}{0.65\linewidth}
    \centering
    \includegraphics[width=.99\linewidth]{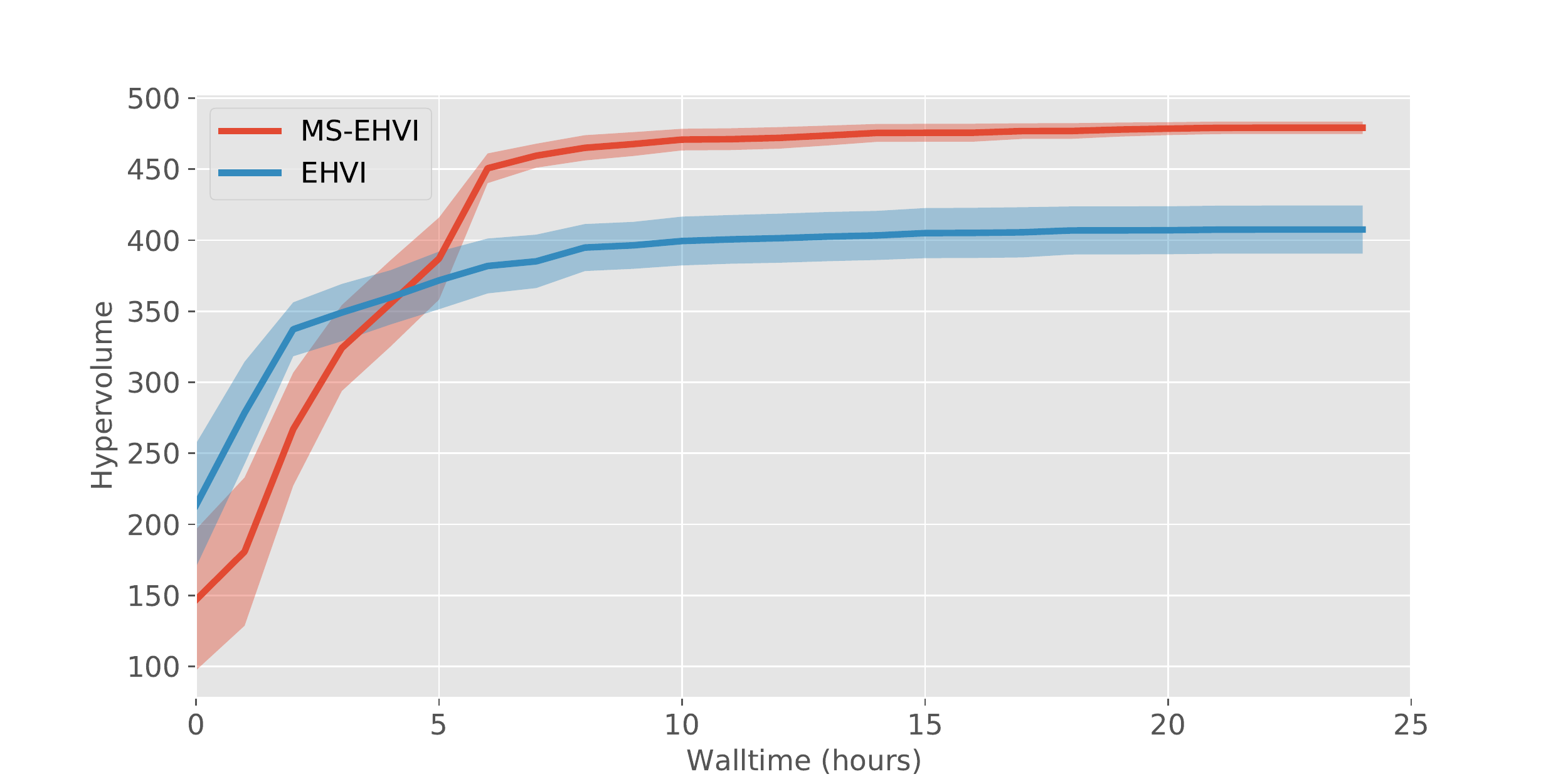}
  \end{subfigure}  
  \caption{Comparison between EHVI and \mdehvi{} on Fashion-MNIST. Hypervolume obtained by both methods over time. We show mean $\pm$ standard error of the mean based on 10 independent runs of each method.}  
  \label{fig:ehvi_basline}
\end{figure}

Figure \ref{fig:ehvi_basline} shows a comparison of \mdehvi{} to standard EHVI.

\section{Supplemental Results}

\begin{figure}
  \centering
  \begin{subfigure}{0.2\linewidth}
    \centering
    \includegraphics[width=.99\linewidth]{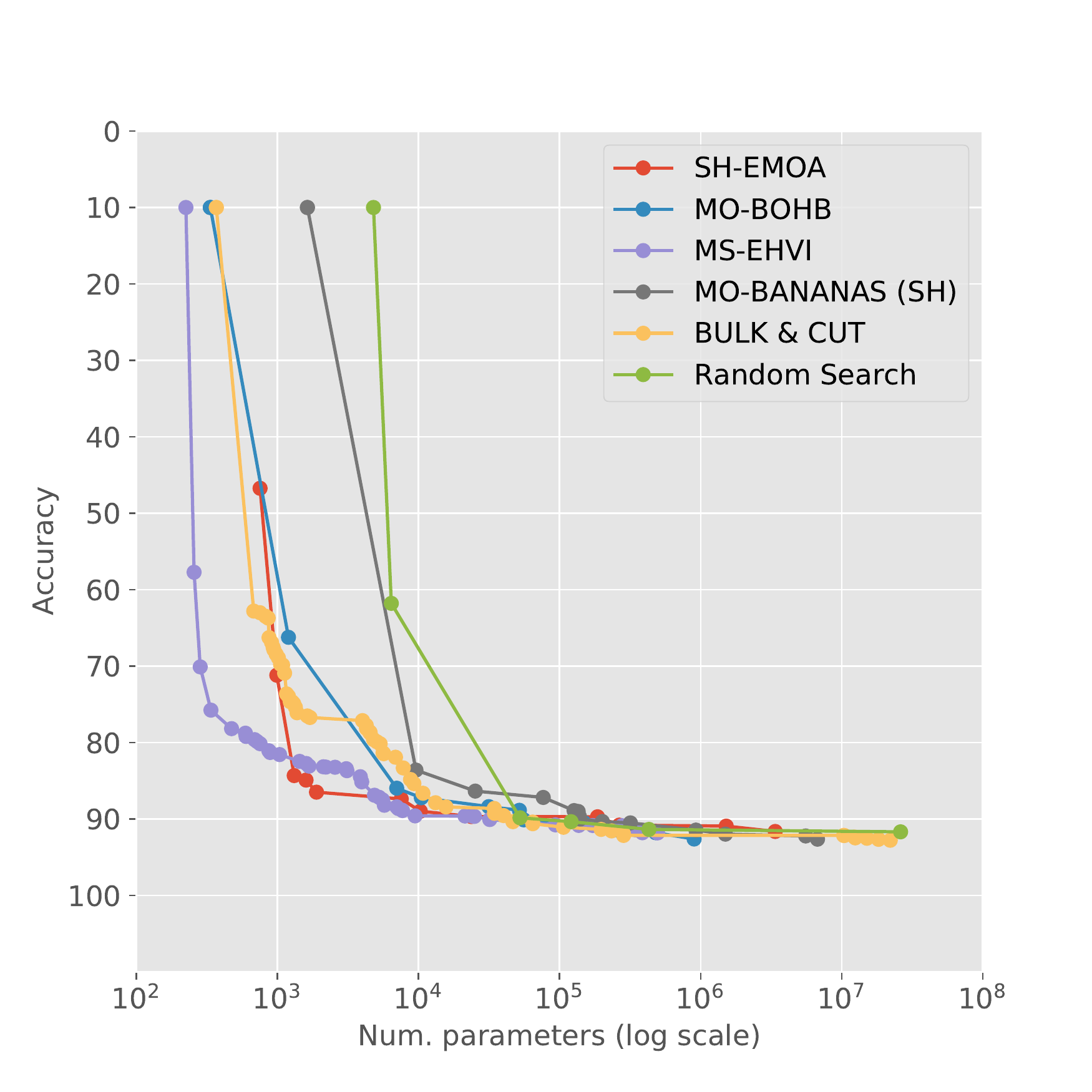}
  \end{subfigure}%
  \begin{subfigure}{0.2\linewidth}
    \centering
    \includegraphics[width=.99\linewidth]{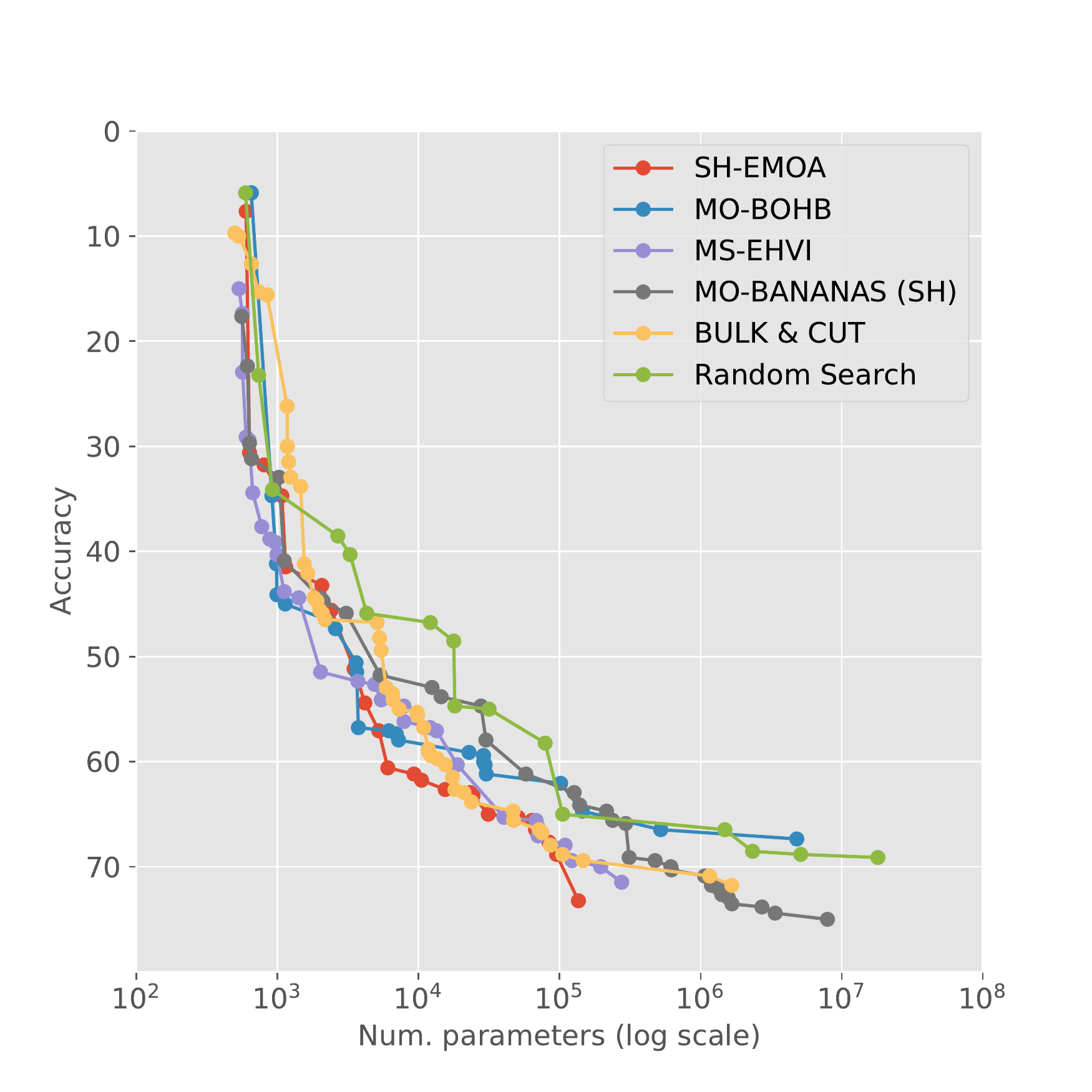}
  \end{subfigure}%
  \begin{subfigure}{0.2\linewidth}
    \centering
    \includegraphics[width=.99\linewidth]{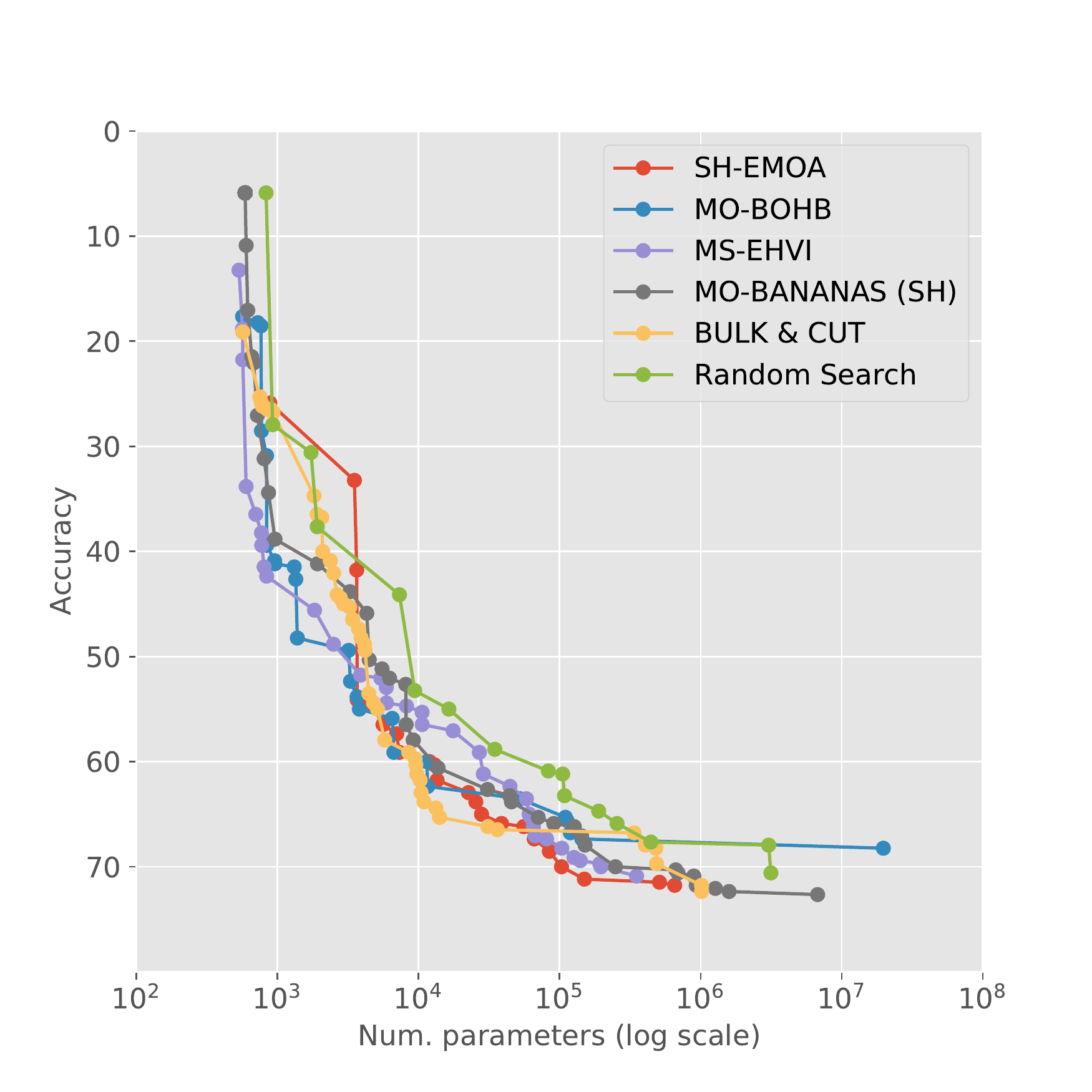}
  \end{subfigure}%
  \begin{subfigure}{0.2\linewidth}
    \centering
    \includegraphics[width=.99\linewidth]{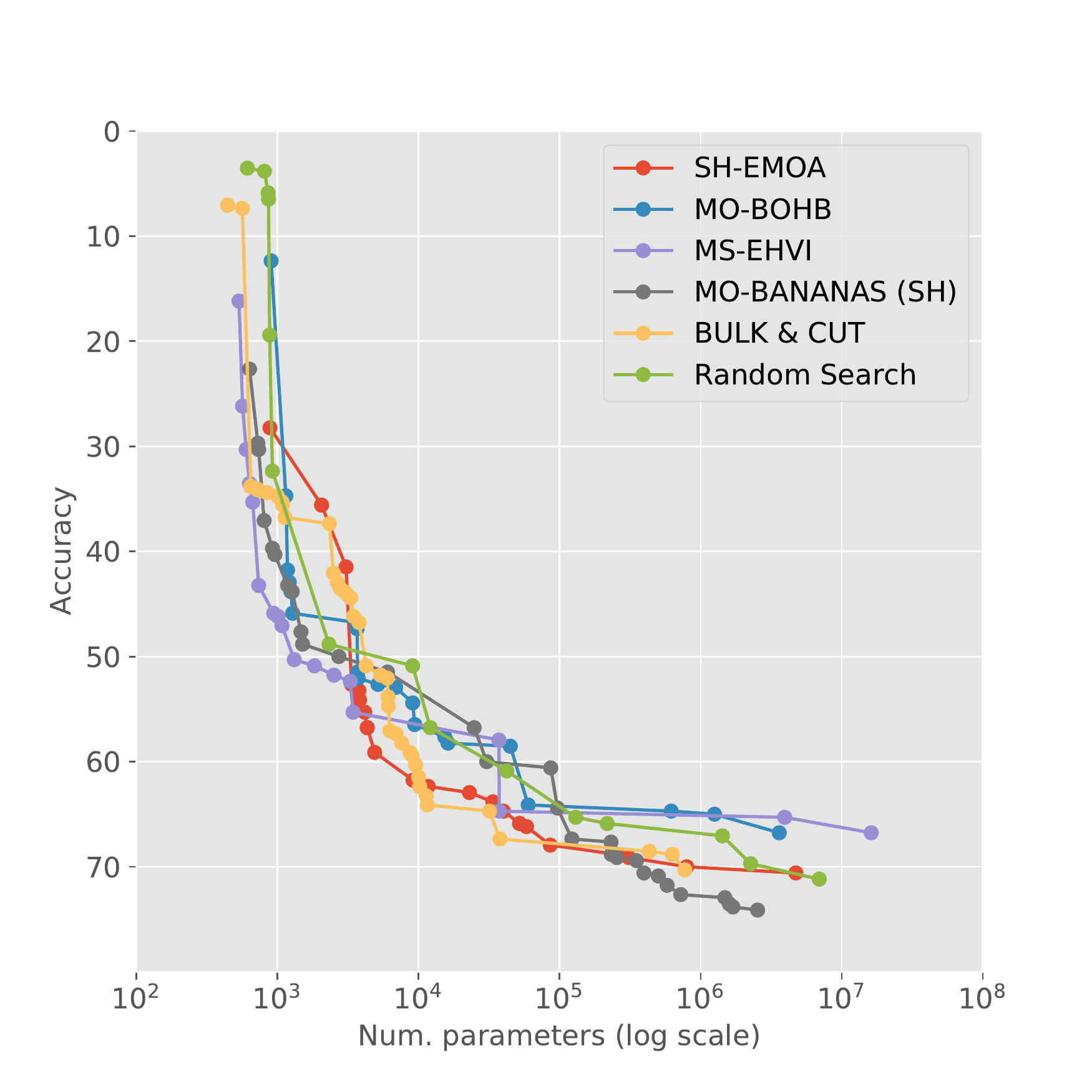}
  \end{subfigure}%
  \begin{subfigure}{0.2\linewidth}
    \centering
    \includegraphics[width=.99\linewidth]{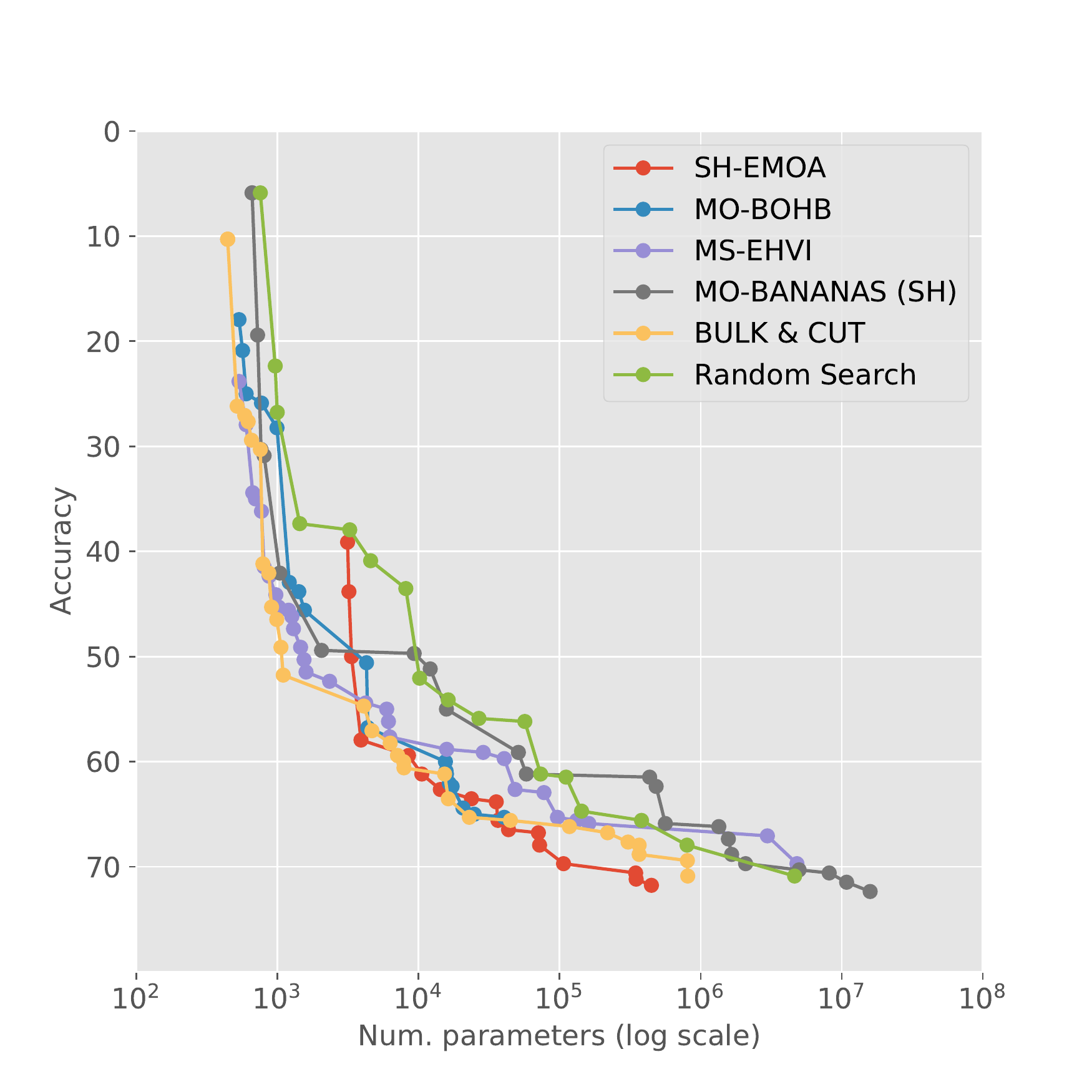}
  \end{subfigure}
  \begin{subfigure}{0.2\linewidth}
    \centering
    \includegraphics[width=.99\linewidth]{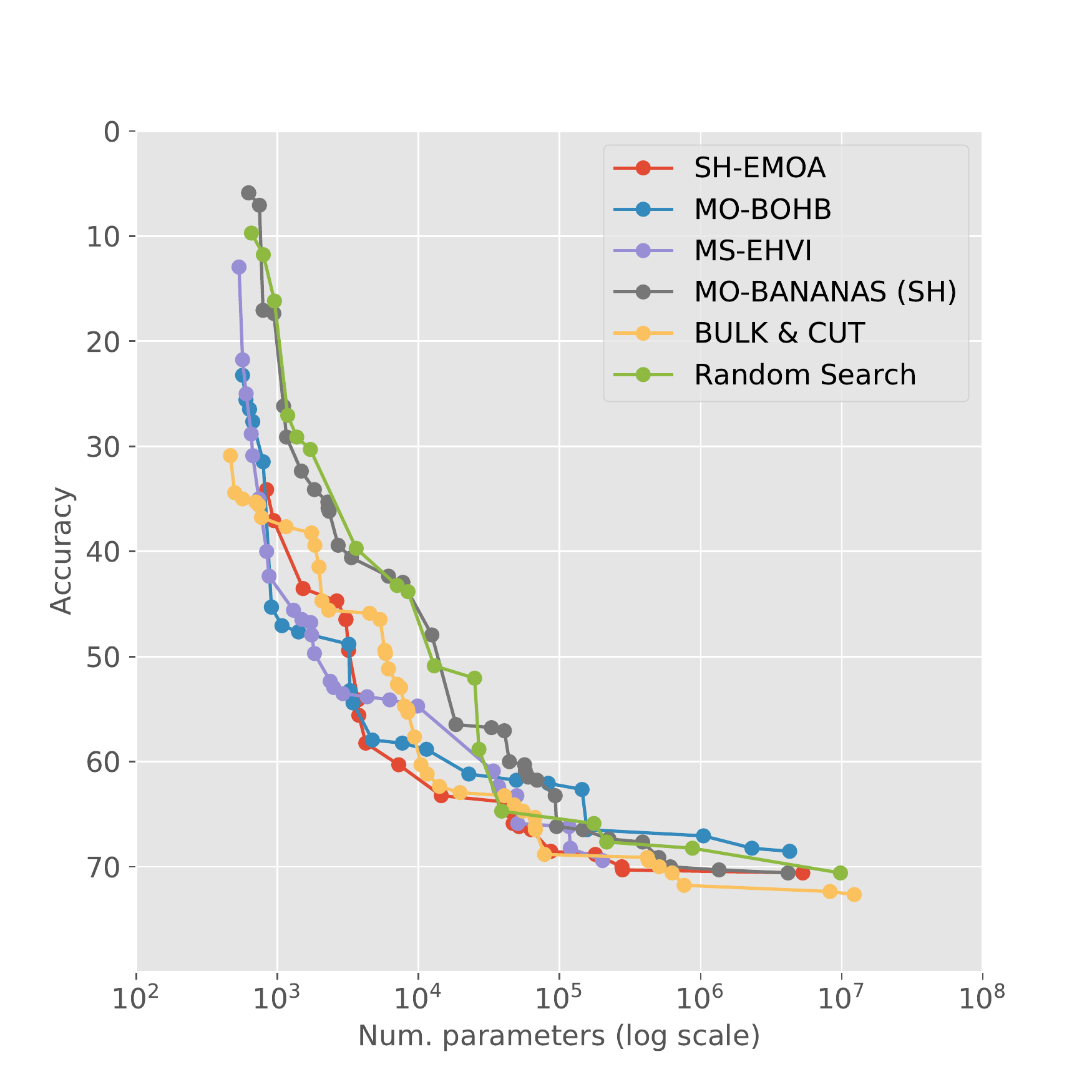}
  \end{subfigure}%
  \begin{subfigure}{0.2\linewidth}
    \centering
    \includegraphics[width=.99\linewidth]{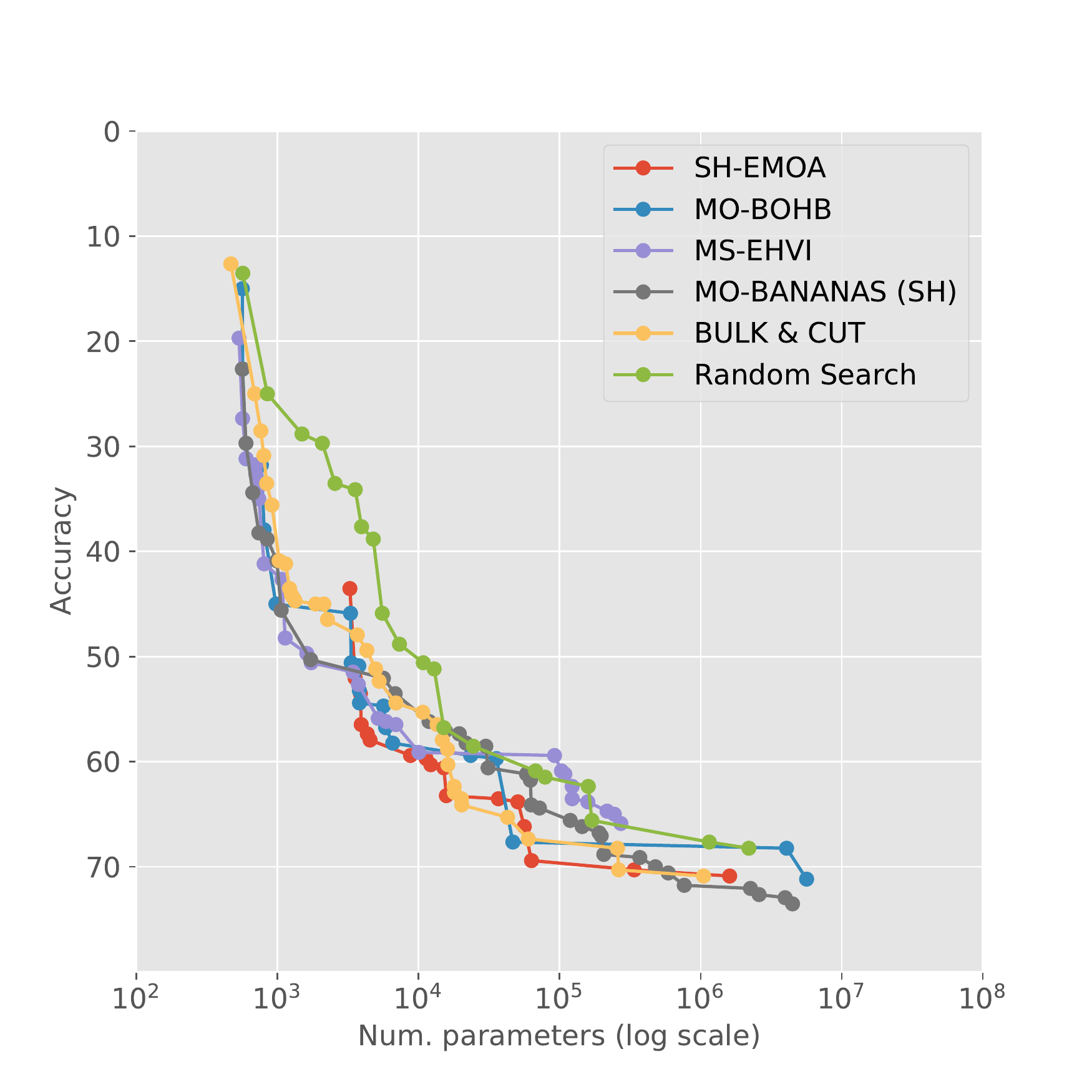}
  \end{subfigure}%
  \begin{subfigure}{0.2\linewidth}
    \centering
    \includegraphics[width=.99\linewidth]{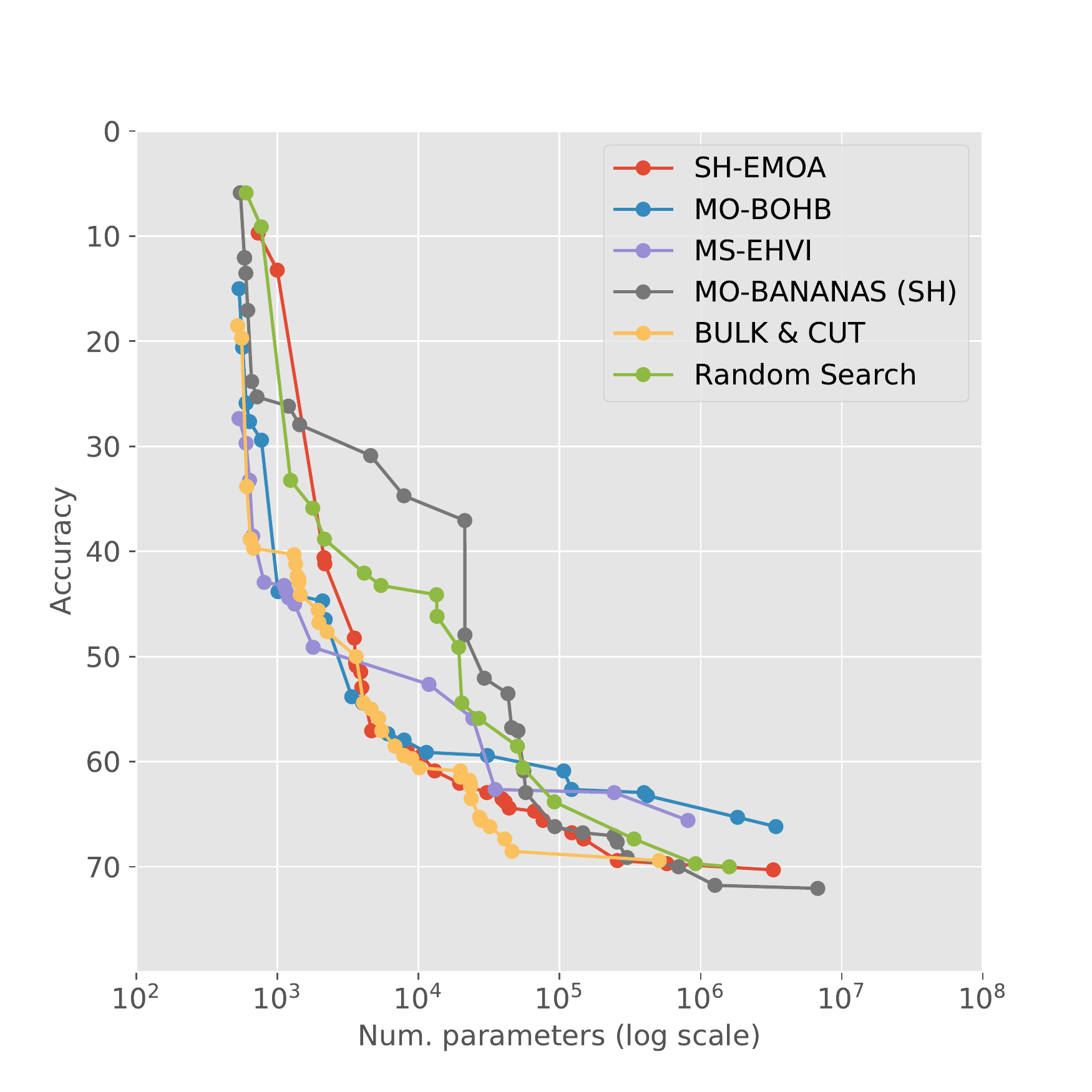}
  \end{subfigure}%
  \begin{subfigure}{0.2\linewidth}
    \centering
    \includegraphics[width=.99\linewidth]{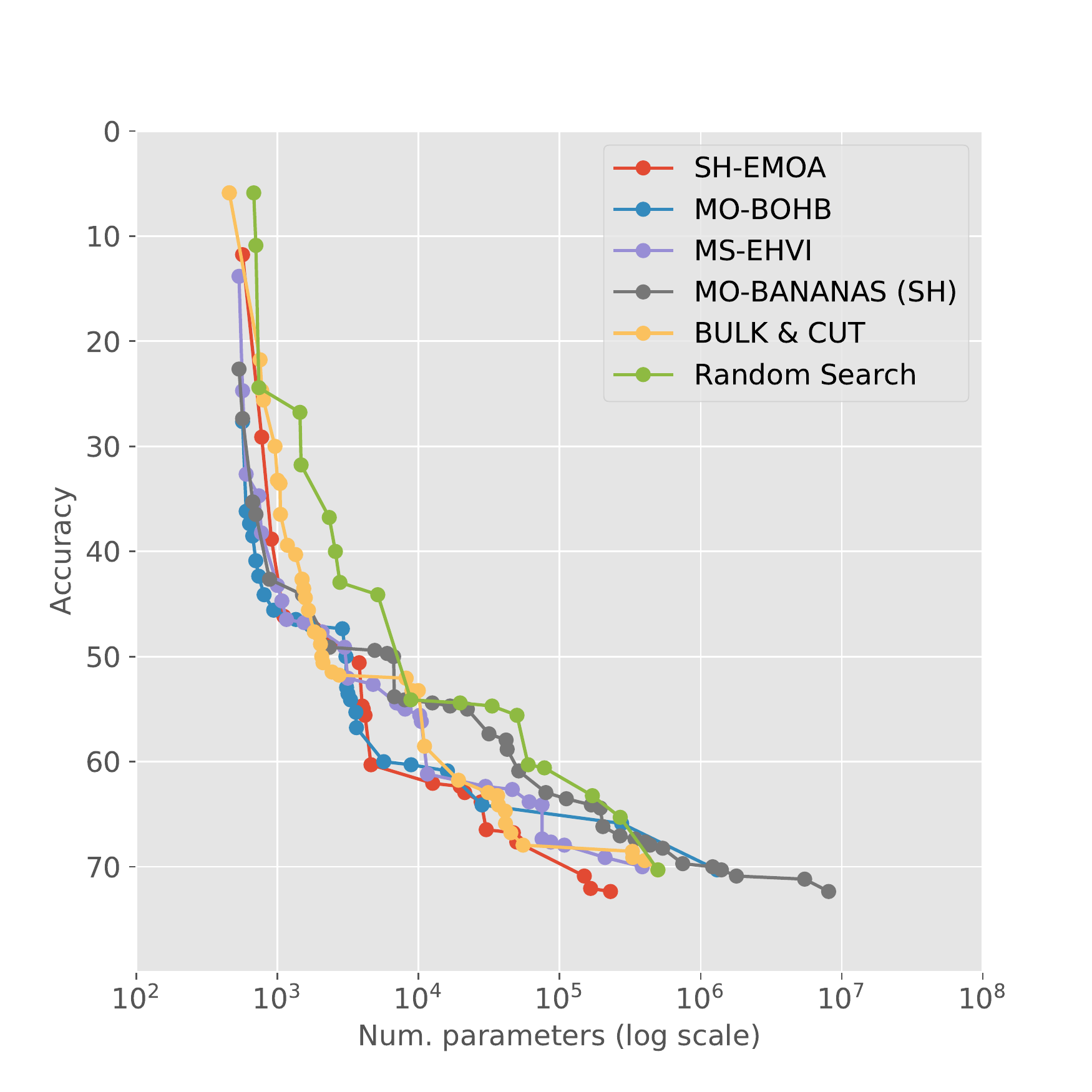}
  \end{subfigure}%
  \begin{subfigure}{0.2\linewidth}
    \centering
    \includegraphics[width=.99\linewidth]{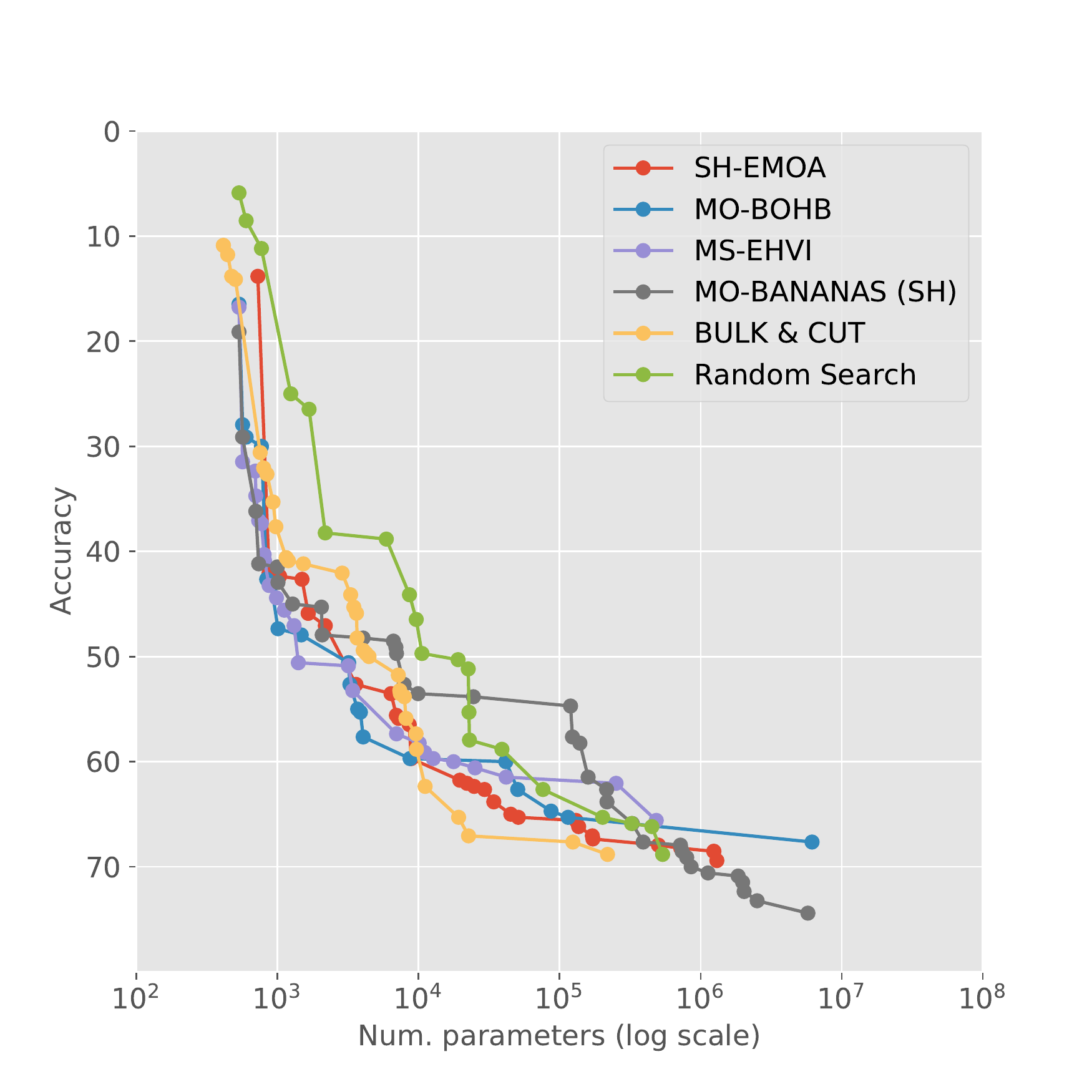}
  \end{subfigure}%
  \caption{Pareto fronts obtained for different initial random seeds on Flowers dataset.}  
  \label{fig:pf_final_app_1}
\end{figure}

\begin{figure}
  \centering
  \begin{subfigure}{0.2\linewidth}
    \centering
    \includegraphics[width=.99\linewidth]{figures/fashion/paretofronts_tst_fashion_0.pdf}
  \end{subfigure}%
  \begin{subfigure}{0.2\linewidth}
    \centering
    \includegraphics[width=.99\linewidth]{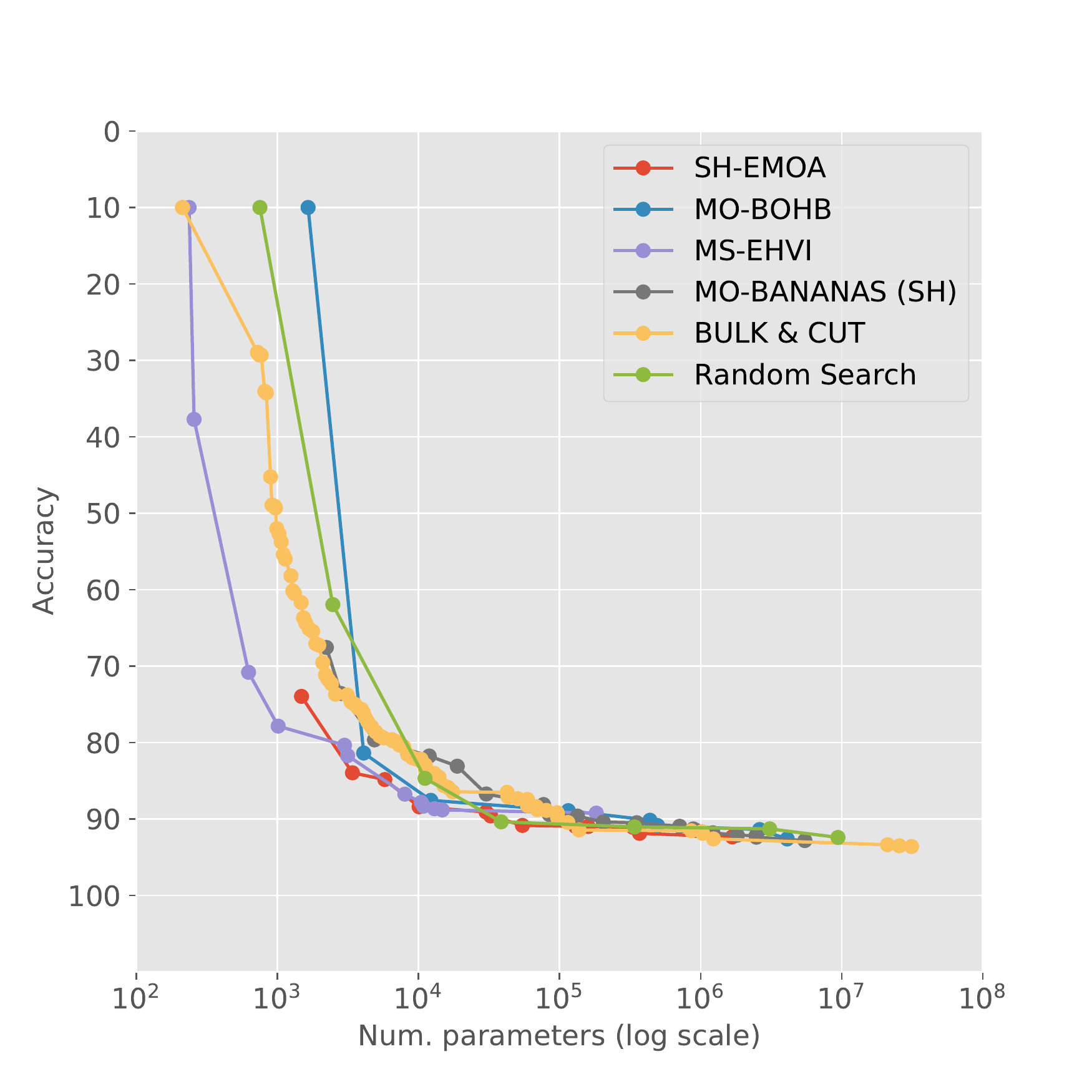}
  \end{subfigure}%
  \begin{subfigure}{0.2\linewidth}
    \centering
    \includegraphics[width=.99\linewidth]{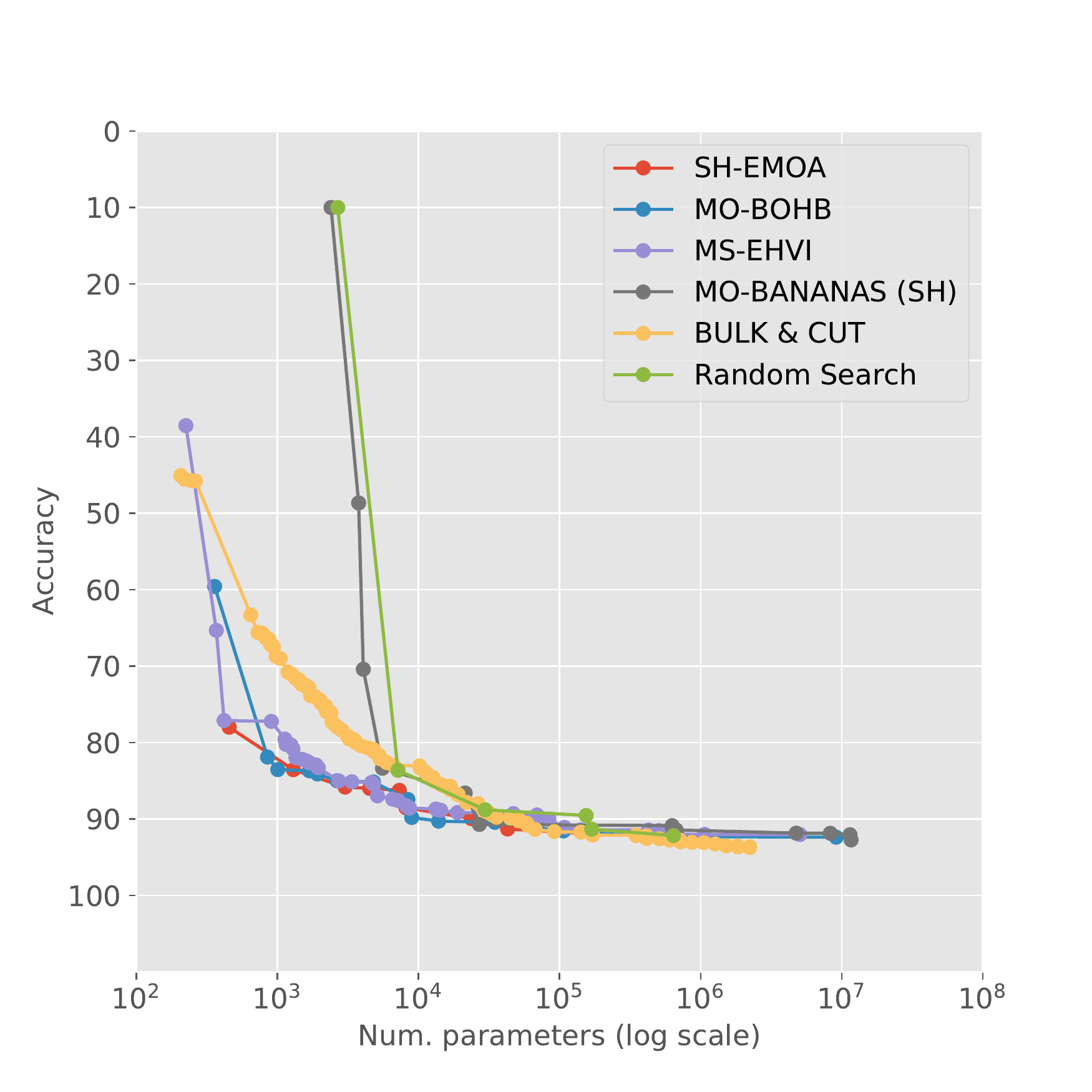}
  \end{subfigure}%
  \begin{subfigure}{0.2\linewidth}
    \centering
    \includegraphics[width=.99\linewidth]{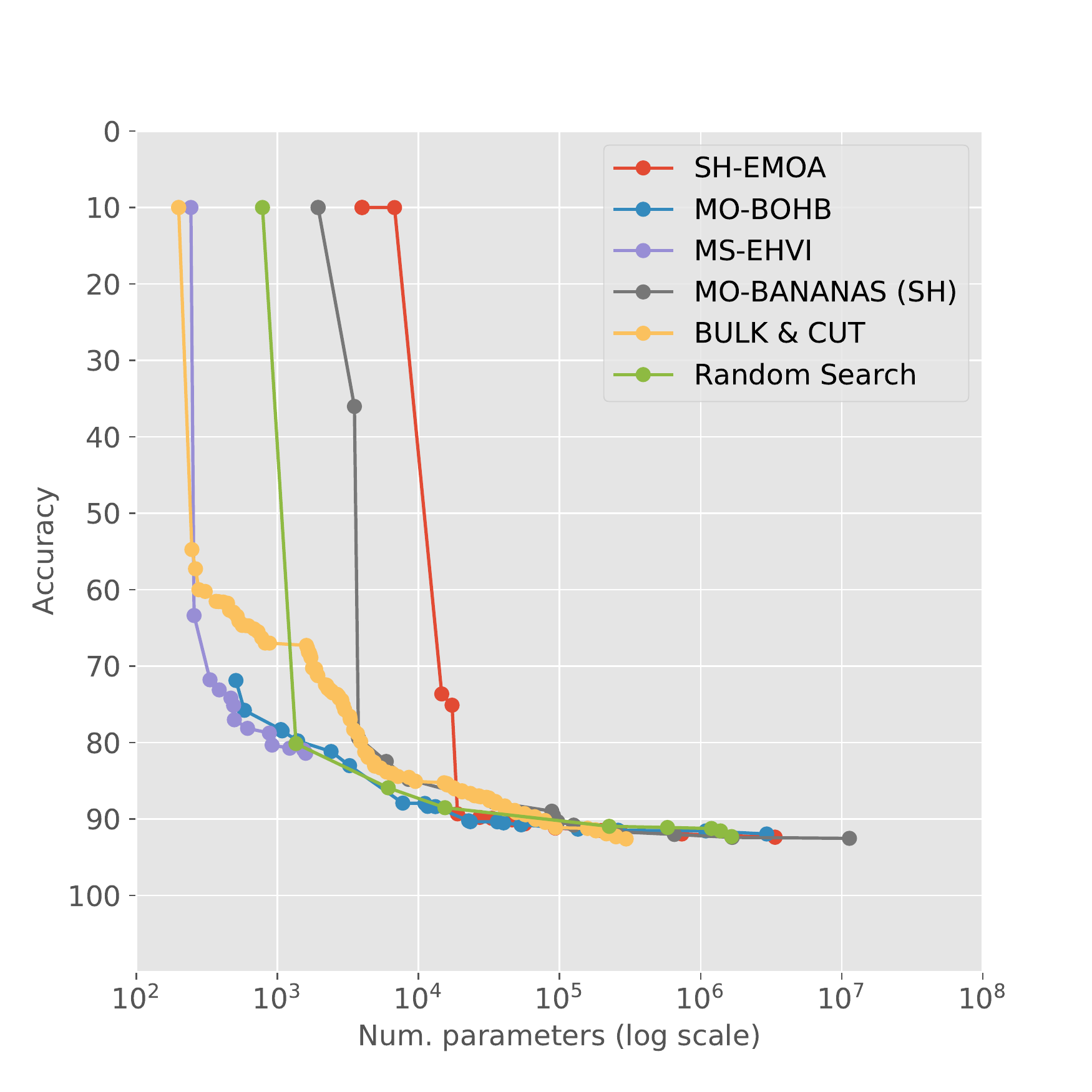}
  \end{subfigure}%
  \begin{subfigure}{0.2\linewidth}
    \centering
    \includegraphics[width=.99\linewidth]{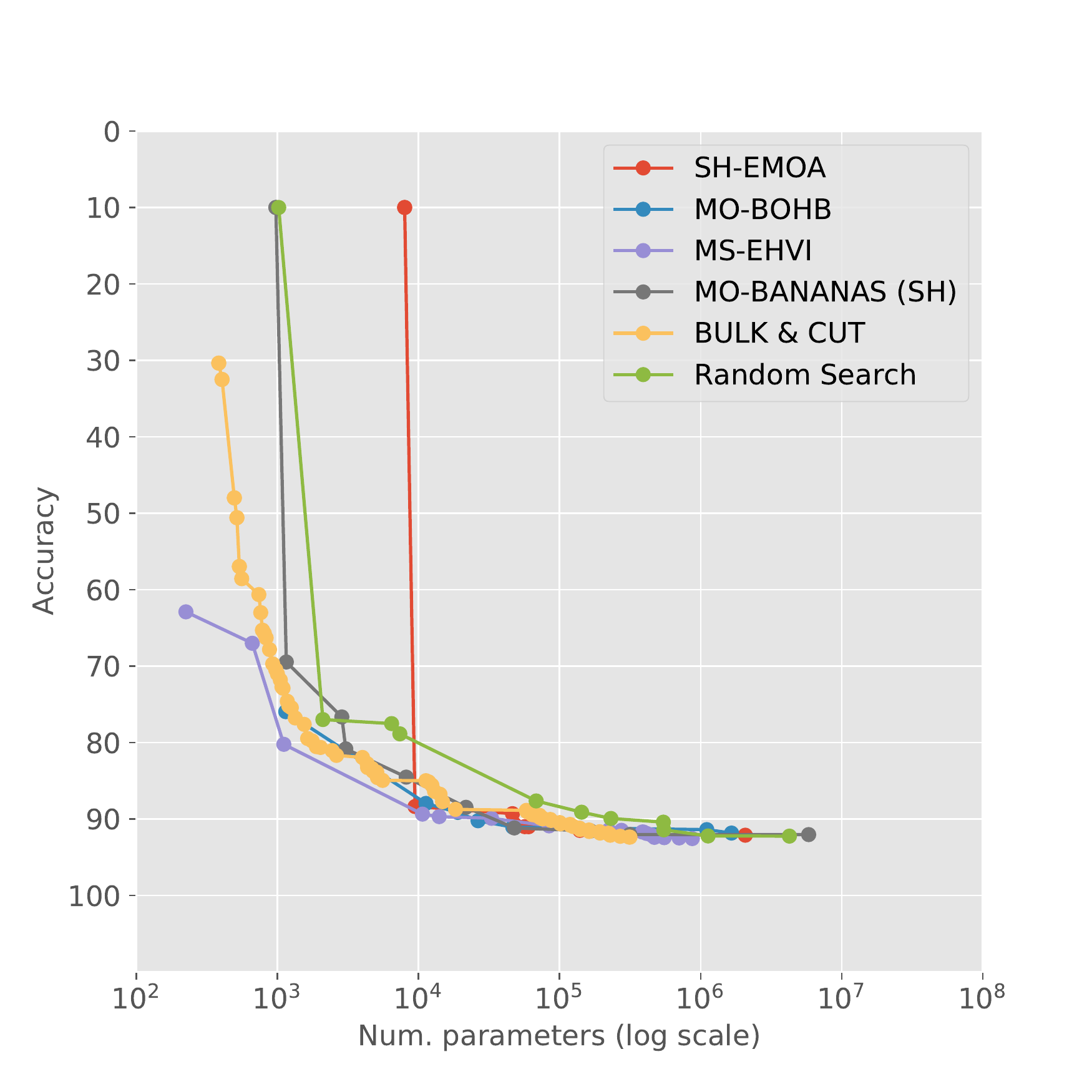}
  \end{subfigure}
  \begin{subfigure}{0.2\linewidth}
    \centering
    \includegraphics[width=.99\linewidth]{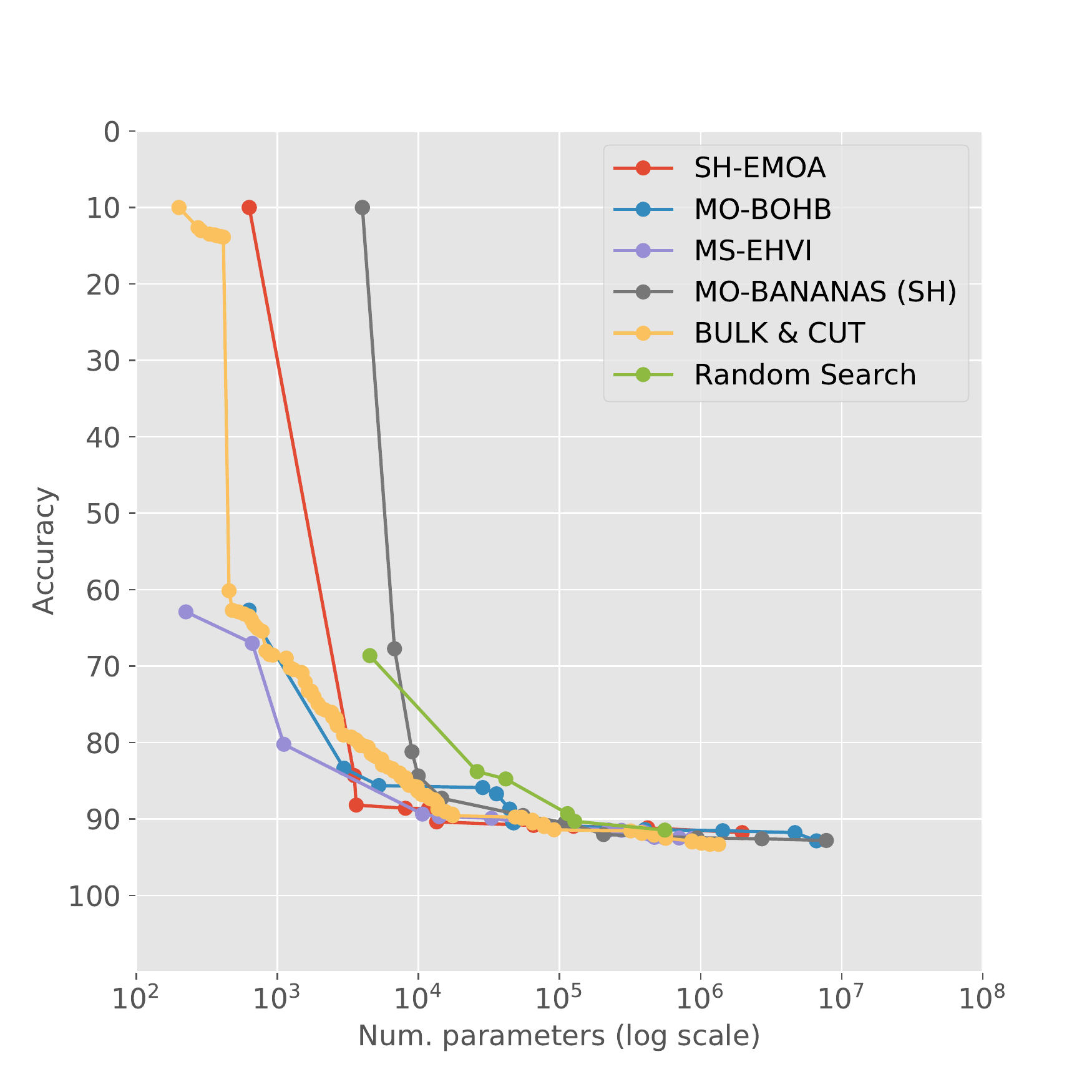}
  \end{subfigure}%
  \begin{subfigure}{0.2\linewidth}
    \centering
    \includegraphics[width=.99\linewidth]{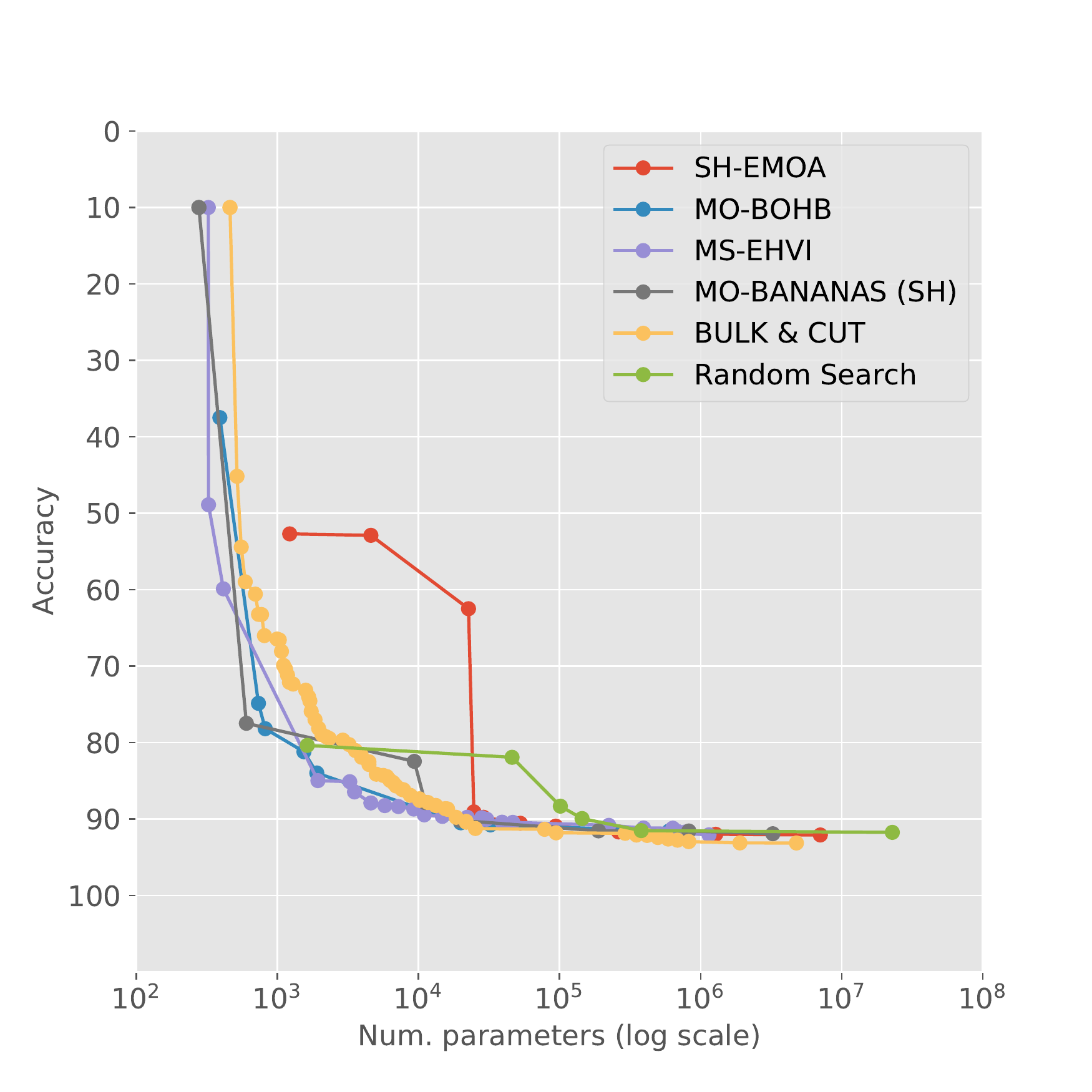}
  \end{subfigure}%
  \begin{subfigure}{0.2\linewidth}
    \centering
    \includegraphics[width=.99\linewidth]{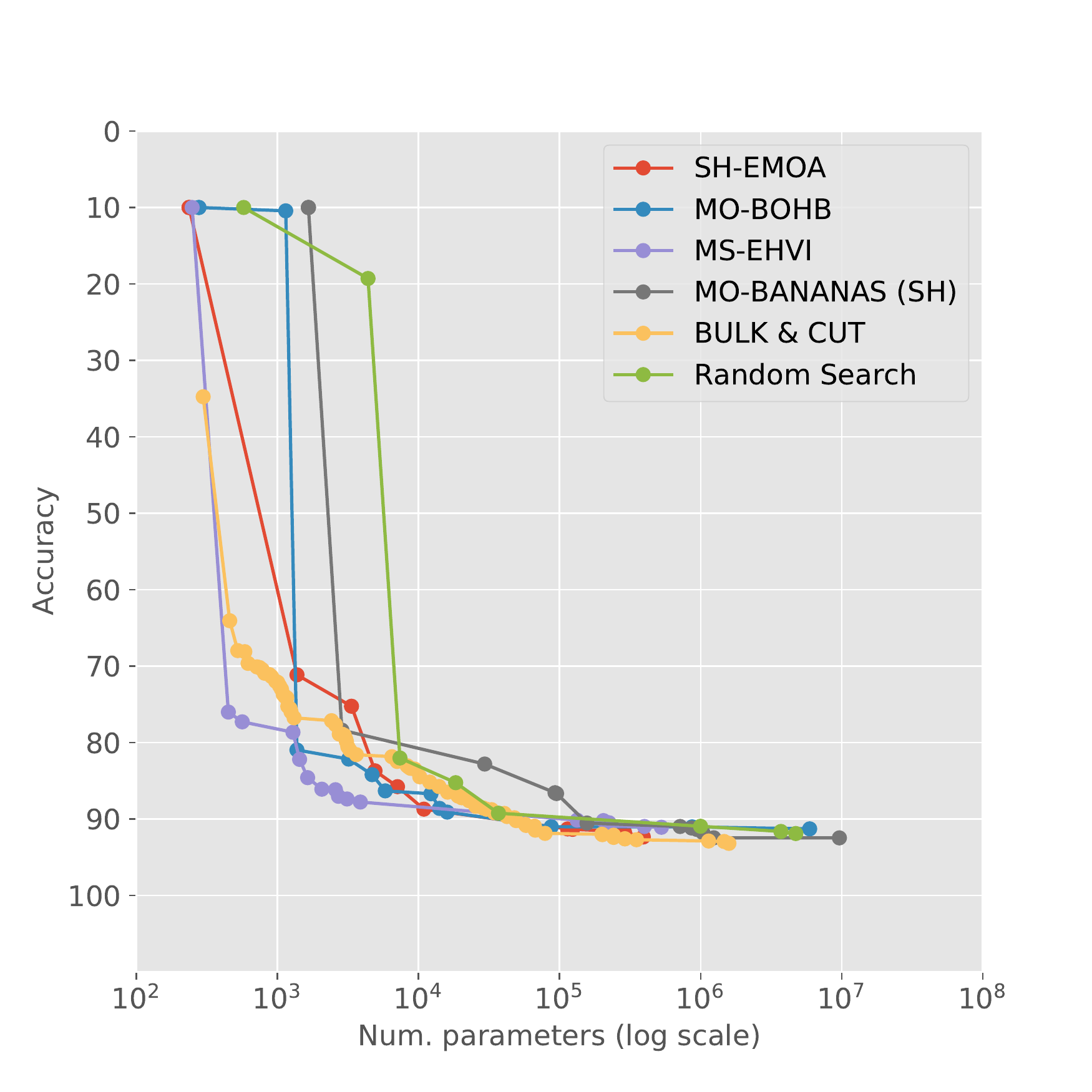}
  \end{subfigure}%
  \begin{subfigure}{0.2\linewidth}
    \centering
    \includegraphics[width=.99\linewidth]{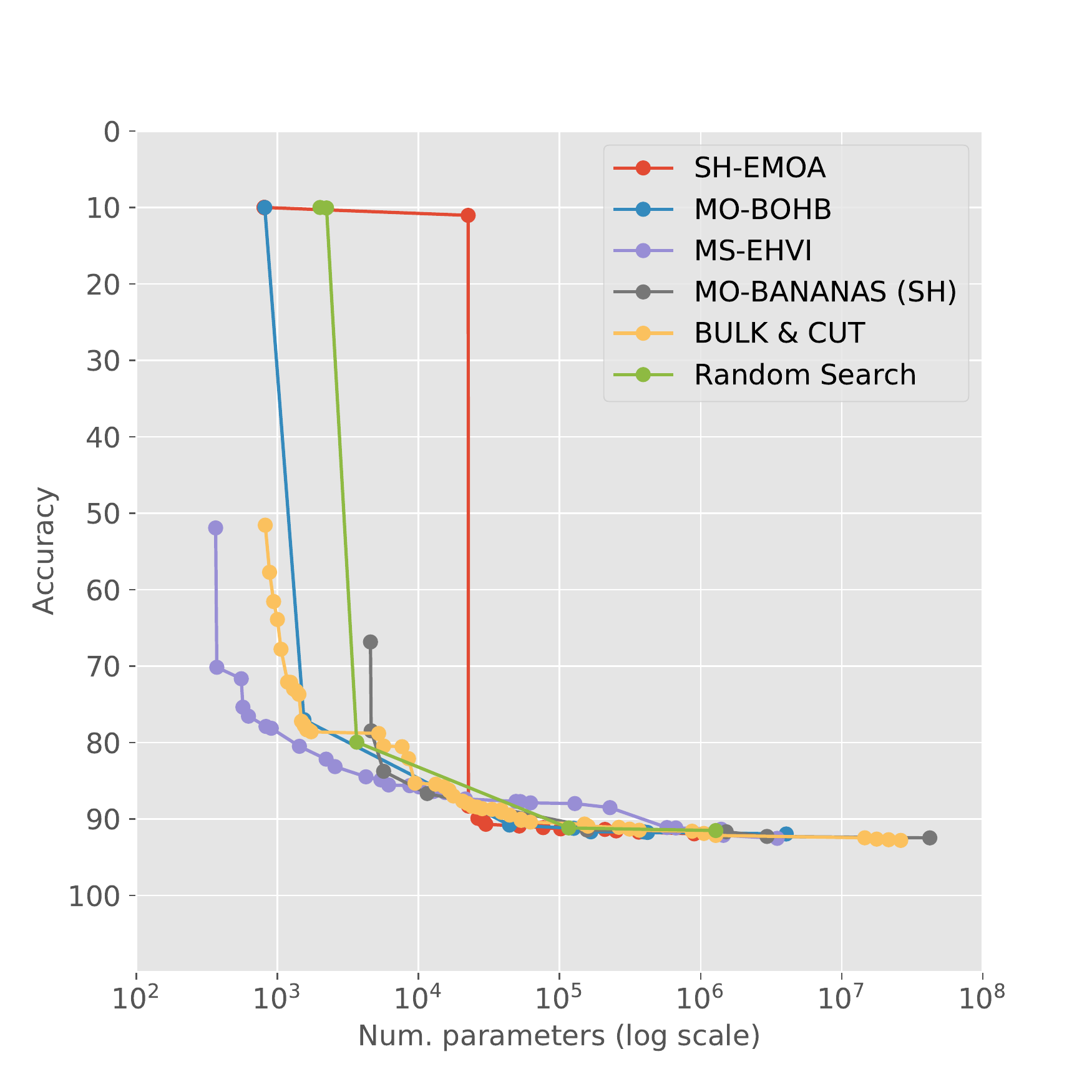}
  \end{subfigure}%
  \begin{subfigure}{0.2\linewidth}
    \centering
    \includegraphics[width=.99\linewidth]{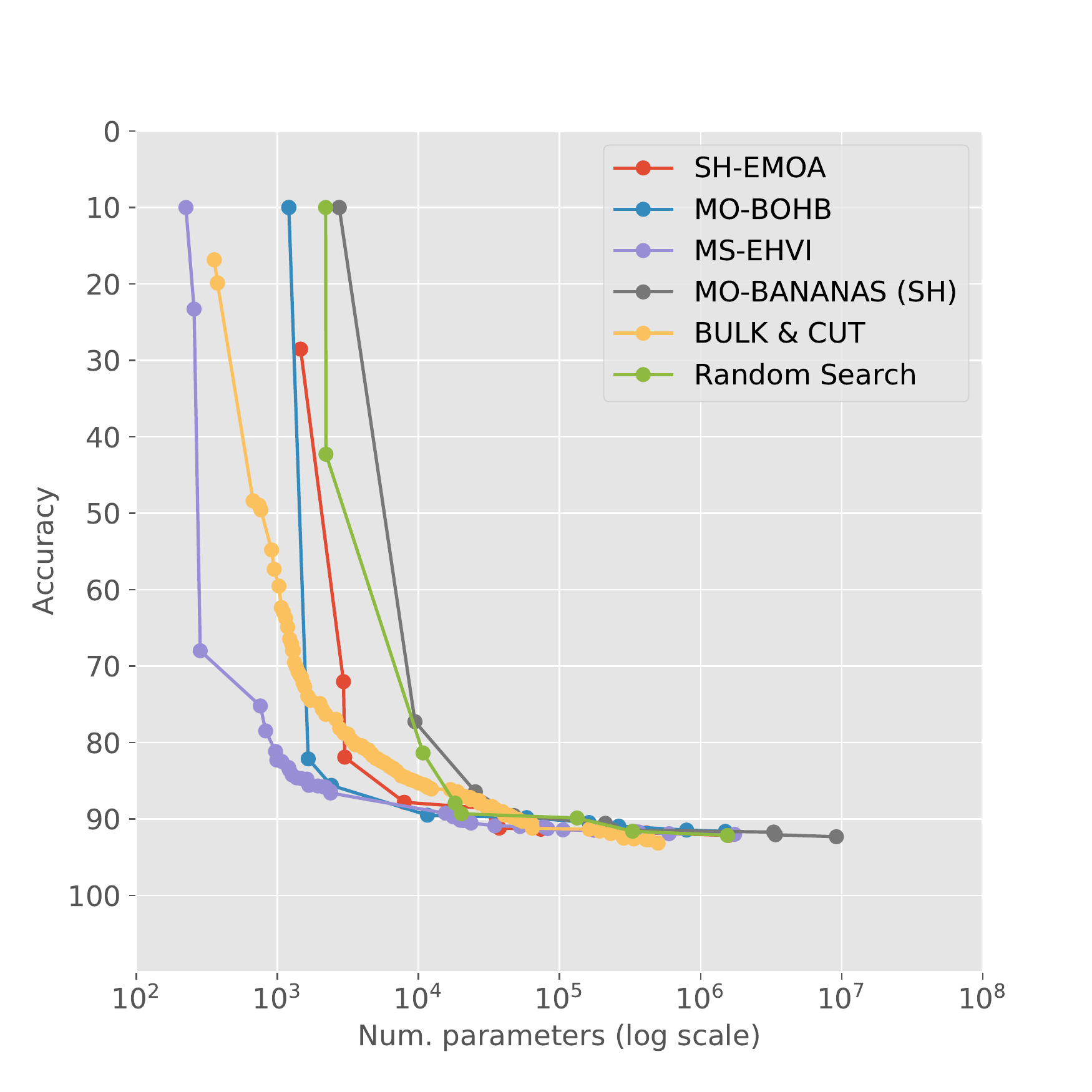}
  \end{subfigure}%
  \caption{Pareto fronts obtained for different initial random seeds on Fashion-MNIST.}  
  \label{fig:pf_final_app_2}
\end{figure}

Figures \ref{fig:pf_final_app_1} and \ref{fig:pf_final_app_2} show all the Pareto fronts found with different seeds.

\end{document}